\documentclass[3p,times]{elsarticle}

\usepackage{ecrc}

\volume{00}

\firstpage{1}

\journalname{Pattern Recognition}

\runauth{}

\jid{procs}

\jnltitlelogo{Pattern Recognition}

\CopyrightLine{2011}{Published by Elsevier Ltd.}

\usepackage{amssymb}

\usepackage[figuresright]{rotating}

\usepackage[super]{cite}
\usepackage{amsfonts}
\usepackage{setspace}
\usepackage{parskip}

\usepackage{graphicx}
\usepackage{caption}
\usepackage{amsmath}
\usepackage{lipsum}
\usepackage{subfig}
\usepackage{algorithm}
\usepackage{url}
\usepackage[table]{xcolor}
\usepackage{seqsplit}
\usepackage{comment}
\usepackage[section]{placeins}

\usepackage{placeins}
\usepackage{algpseudocode}
\urldef{\footurl}\url{https://github.com/aralab-unr/CoSM-ICP}

\begin{document}

\begin{frontmatter}



\dochead{}

\title{Registration of 3D Point Sets Using Correntropy Similarity Matrix}


\author{Ashutosh Singandhupe$\dagger$, Hung Manh La$\dagger$ \thanks{Corresponding author: Hung La, E-mail: hla@unr.edu}, Trung Dung Ngo, Van Ho}
\address{Ashutosh Singandhupe and Dr. Hung La are with the Advanced Robotics and Automation
(ARA) Laboratory, Department of Computer Science and Engineering, University of Nevada, Reno, NV 89557, USA} 
\address{Dr. Trung Dung Ngo is with the More-Than-One
Robotics Laboratory, University of Prince Edward Island, Canada}
\address{Dr. Van Ho is with the Soft Haptics Laboratory, School of Materials Science,
Japan Advanced Institute of Science and Technology (JAIST), Japan}

\begin{abstract}
This work focuses on Registration or Alignment of 3D point sets. Although the Registration problem is a well established problem and it's solved using multiple variants of Iterative Closest Point (ICP) Algorithm, most of the approaches in the current state of the art still suffers from misalignment when the  \textit{Source} and the \textit{Target} point sets are separated by large rotations and translation. In this work, we propose a variant of the Standard ICP algorithm, where we introduce a Correntropy Relationship Matrix in the computation of rotation and translation component which attempts to solve the large rotation and translation problem between \textit{Source} and \textit{Target} point sets. This matrix is created through correntropy criterion which is updated in every iteration. The correntropy criterion defined in this approach maintains the relationship between the points in the \textit{Source} dataset and the \textit{Target} dataset.  Through our experiments and validation we verify that our approach has performed well under various rotation and translation in comparison to the other well-known state of the art methods available in the Point Cloud Library (PCL) as well as other methods available as open source. We have uploaded our code in the github repository for the readers to validate and verify our approach {\footurl}.
\end{abstract}

\begin{keyword}
Registration, 3D point clound, Correntropy criterion, Iterative Closest Point

\end{keyword}

\end{frontmatter}



\section{Introduction} \label{sec:Intro}
Registering two point clouds is an active area of research in the community. Registration is the process of aligning two point cloud datasets of the same scene, which are separated by a certain transformation (rotation and translation). The problem of registration can be explained as finding the transformation between two point clouds such that the relative error between the 2 point sets is minimized. The problem of registration is a significant component in Simultaneous Localization and Mapping (SLAM) for robotics and other visual data matching algorithms \cite{AshuIRC2018}. The relative transformation computed from point clouds acquired at close time intervals from a 3D sensor can be used to compute the overall trajectory of a system \cite{Cadena_IRO2016,AshuIRC2018}.

Early approaches for 3D point cloud registration can be attributed to Besl and MacKay \cite{10.1109/TPAMI.1987.4767965,121791} and Arun et.al~\cite{10.1109/TPAMI.1987.4767965}, which consists of computing the centroid for both \textit{Source} and \textit{Target} datasets. Each point from both the datasets is subtracted from their respective centroids and later on, Singular Value Decomposition (SVD) is used to compute the rotation component. This procedure is performed iteratively until the relative mean square error between the point cloud datasets is minimized, and hence it is termed as Iterative Closest Point (ICP) Algorithm. Since then variants of ICP have been introduced, a point-to-plane variant of ICP operates on taking advantage of surface normal information \cite{articleptp}. Instead of minimizing the root mean square of the individual points as done in the generalized version of the ICP, the point to plane algorithm minimizes error along the surface normal~\cite{Zhang2015ICRA,tang2018learning,Tateno2017CNNSLAMRD,SehgalISVC2019,choy2020deep,Yang2020TEASERFA,inproceedingszhou}.


Despite the various advances in the registration methodologies, handling the situation where the data set is noisy is still a major concern. If the noise is Gaussian, we can remove most of it from the data. However, when the noise is non-Gaussian (e.g., shot noises), it becomes difficult to resolve. Hence, it is important to address the problem of handling non-Gaussian noise. The idea of Correntropy has seen widespread use in modern machine learning, signal processing, and other related fields  \cite{Ashu_MCCEKFSec, WU2019PR}. Correntropy is a measure of similarity between two random variables and has been implemented with the traditional Kalman filter   \cite{10.1115/1.4043054}. Some recent work \cite{WU2019PR, zhang2018robust, xu2018precise,articlezhang,8500525} has introduced the Correntropy concept to the registration problem to overcome the drawbacks of the ICP algorithm. In the traditional ICP algorithm, if any of the point cloud dataset (\textit{Source} or \textit{Target}) is affected by non-Gaussian noise, the ICP fails drastically and affects the overall transformation estimation of the point clouds. In this work, we show through basic examples on how the addition of non-Gaussian noise can affect the traditional ICP as well as compare how our approach can be better than the traditional ICP algorithm and it's variants. We also compare our approach with the well known  Normal Distribution Transform (NDT) algorithm available in the PCL library.

In this work, we propose a Correntropy Similarity Matrix Iterative Closest Point (CoSM ICP) algorithm. In our approach, we introduce a matrix of size $N\times N$ which maintains a `numeric' relationship between  \textit{Source} and \textit{Target} dataset points (size of \textit{Source} dataset and \textit{Target} dataset is $N$, hence the matrix size is $N\times N$). This numeric relationship is calculated through the well-known correntropy criterion,  and it is updated at every iteration. It is calculated only between the nearest points from \textit{Source} dataset to the \textit{Target} dataset, thereby making the Similarity Matrix a sparse matrix. Our approach not only provides a significant improvement in the alignment of \textit{Source} and \textit{Target} dataset under various rotations and translations, it also proves quite effective against non-Gaussian outliers affecting the dataset. In addition to evaluating our approach in multiple datasets, we also evaluate the method where we inject the \textit{Source} dataset with random false values and through experimental comparison with the traditional ICP and other approaches, we evaluate how well our approach can estimate the transformation between the 'infected' \textit{Source} and the \textit{Target} dataset and we see how our approach performs better than the well known state of the art methods. Along side estimating accurate transformations, handling shot-noise or the non-Gaussian noise vectors is a critical component for better transformation estimation and is the driving force for our work.
 
 Inspired by this Correntropy concept, we fuse it with the ICP algorithm, called Correntropy Similarity Matrix ICP or CoSM-ICP. Our main contribution to this work can be summarized as follows:
 \begin{itemize}
 \item We propose a new data registration algorithm in which the Correntropy Similarity Matrix ICP is introduced.
 \item We evaluate our approach on various randomly generated transformation matrices between \textit{Source} and the \textit{Target}.
 \item Through evaluation on these random transformation matrices we see how well our approach performs better than the other well known state of the art methods.
 \item We also introduce outliers  in the \textit{Source} dataset  and evaluate our approach on the 'infected' \textit{Source} dataset.
 \item We compare our approach with other approaches on the 'infected' \textit{Source} dataset and see through results how well our approach performs better.
\end{itemize}

%

  The remaining paper is organized as follows: Section \ref{Section_2} introduces the idea of Correntropy and its underlying concepts. Section \ref{Section_3} describes our proposed method and it's implementation. The results and evaluation of our proposed method are discussed in Section \ref{Results}. Conclusions and future work discussion are given in Section V.

\section{Correntropy Criterion}
\label{Section_2} 

\begin{table*}[t!]
\centering 
\caption{Nomenclature}
\label{Tab:Var1}
\begin{tabular}{ll} 
\hline
 Variables & Usages\\
\hline
 $\bold{{P}_s}$ & \textit{Source} Point Cloud Set.\\
 $\bold{{P}_t}$ & $\textit{Target}$ Point Cloud Set.\\
 $N$ & Total number of points in point clouds $\bold{{P}_s}$ and $\bold{{P}_t}$.\\
 $j,k$ & Index of every point in point cloud where $j,k=1,..,N$.\\
 $\bold{p}_j$ & Individual 3D point $(x_j, y_j, z_j)$ in a point cloud $\bold{{P}_s}$, or $\bold{p}_j \in \bold{P}_s$.\\
 $\bold{q}_k$ & Individual 3D point  $(x_k, y_k, z_k)$ in a point cloud $\bold{{P}_t}$, or  $\bold{q}_k \in \bold{P}_t$.\\
 $G_\sigma$ & Gaussian kernel function with bandwidth $\sigma$.\\
 $\bold{R}$ & Rotation matrix.\\
 $\bold{tr}$ & Translation vector.\\
 $\varepsilon^2$ & Root mean square error.\\
 $s$ & Scaling factor.\\
 $\textbf{SM}$ & Similarity Matrix.\\
 $\bold{s_{cen}}$,$\bold{t_{cen}}$ & Centroids computation of \textit{Source} and \textit{Target} datasets.\\
 
\hline
\end{tabular}
\end{table*}

\begin{figure*}[h!]
\centering
\subfloat[]{\includegraphics[width=0.45\columnwidth,height=6cm]{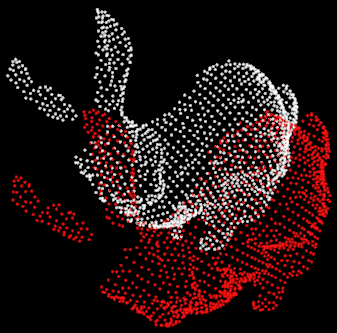}} \hfil
\subfloat[]{\includegraphics[width=0.45\columnwidth,height=6cm]{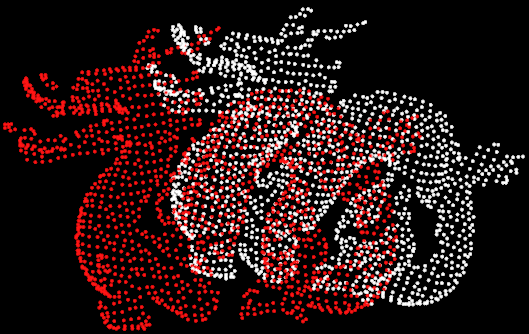}}
\caption{(a) Red is the \textit{Source} point cloud $\bold{{P}_s}$ (bunny rabbit). $\bold{{P}_s}$ contains $N$ points $\bold{p}_j$ ($j=1,..,N$), each of which is a 3D representation
  $x_j,y_j,z_j$. White is the \textit{Target} point cloud $\bold{{P}_t}$. $\bold{{P}_t}$ contains $N$ points $\bold{q_{k}}$ ($k=1,..,N$), each of which is a 3D representation
  $x_k,y_k,z_k$. Same goes for the dragon dataset in (b).}
\label{fig:initbunny1}
\end{figure*}

We begin by describing the mathematical representation of point cloud data and develop our algorithm based on it. 
We denote $\textit{Source}$ point cloud as $\bold{{P}_s}$ and $\textit{Target}$ point cloud as $\bold{{P}_t}$.  The points in the point cloud $\bold{{P}_s}$ contains $N$ points in which each point can be referenced as $\bold{p}_j=\left \{x_j,y_j,z_j  \right \}$, where $x_j,y_j,z_j$ denotes the 3D coordinates of the point $\bold{p}_j$. Similarly, the points in the point cloud $\bold{{P}_t}$ contains $N$ points in which each point can be represented as $\bold{q}_k=\left \{x_k,y_k,z_k \right \}$, where $x_k,y_k,z_k$ denotes the 3D coordinates of the point $\bold{q}_k$. Throughout the paper, we describe a matrix as bold and capitalized letters, a vector is represented as bold and lowercase letters and the rest of the variables (single dimension) are represented as lowercase variables.
Figure~\ref{fig:initbunny1} shows a brief description of the point clouds and Table~\ref{Tab:Var1} denotes the mathematical notations used throughout this work.

We use the Correntropy concept between the \textit{Source} point cloud and the \textit{Target} point cloud. Correntropy has proved beneficial to remove large outliers \cite{Ashu_MCCEKFSec}. Correntropy is essentially a similarity measure of two random variables. We use the Correntropy concept between \textit{Source} and \textit{Target} point clouds. In this scenario, we are measuring the similarity between points $\bold{p}_j$ and $\bold{q}_k$ from the point clouds $\bold{{P}_s}$ and $\bold{{P}_t}$, respectively. Then the Correntropy criterion between points $\bold{p}_j$ and $\bold{q}_k$ can be mathematically represented as :
\begin{equation}
    \bold{\textit{V}_\sigma} (\bold{p}_j,\bold{q}_k)= \bold{\textit{E}}[\kappa _\sigma(\bold{p}_j-\bold{q}_k)].
    \label{Correntropy}
\end{equation}

Equ. (\ref{Correntropy}) also refers to in common literature as a cross-Correntropy of two random variables~\cite{conf/ciss/IzanlooFYS16,CHEN201770,10.1115/1.4043054}. 
 In Equ. (\ref{Correntropy}), $\bold{\textit{E[.]}}$ refers to the expectation of the variable, and $\kappa_\sigma$ denotes the kernel function. We can manually choose the size of the kernel function (also referred to as bandwidth $\sigma$). In our approach, we use the Gaussian kernel function where we redefine the Correntropy function as:
\begin{equation}
   \bold{\textit{V}_\sigma}(\bold{p}_j,\bold{q}_k)=\frac{1}{N}\sum_{j,k=1}^{N}\bold{\textit{G}_\sigma}(\bold{p}_j-\bold{q}_k).
   \label{G-Correntropy}
\end{equation}
From Equ.(\ref{G-Correntropy}),
\begin{equation}
    G_\sigma(\bold{p}_j-\bold{q}_k)=exp(-\frac{\left \| \bold{p}_j-\bold{q}_k \right \|^2}{2\sigma^2}),
    \label{Gsigma}
\end{equation} 
where $\sigma$ is the bandwidth or the kernel size of the Gaussian kernel. From Equ. (\ref{Gsigma}) it becomes evident that if $\bold{p}_j=\bold{q}_k$ Gaussian Correntropy is maximum, and the Gaussian Correntropy function is positive and bounded~\cite{conf/ciss/IzanlooFYS16,CHEN201770,10.1115/1.4043054}. 

\begin{figure}[htb]
\centering
    \includegraphics[width=0.5\columnwidth,height=9cm]{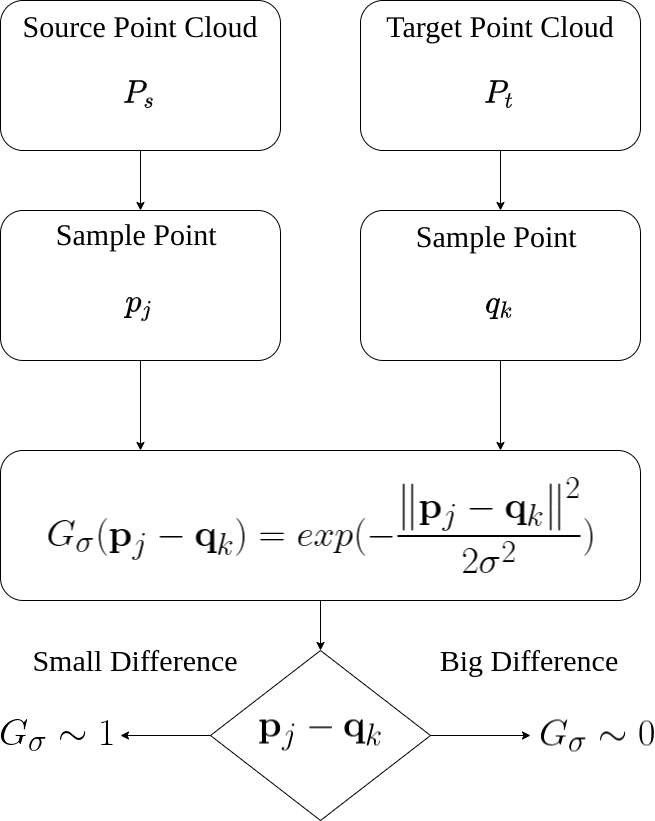}
\caption{Correntropy Criterion. Here, when the difference between the points ($\bold{p}_j-\bold{q}_k$) is small, i..e., when they are similar, and the Gaussian kernel function $G_\sigma$ approaches to 1. On the contrary, when the difference is large, the Gaussian kernel function approaches to 0.  }
\label{Correntropy_Criterion}
\end{figure}

Fig. \ref{Correntropy_Criterion} shows the basic understanding of the Correntropy criterion. We see that when the difference between the points ($\bold{p}_j-\bold{q}_k$) is small, \textit{i.e.,} they are similar, and the Gaussian kernel function $G_\sigma$ approaches to 1. On the contrary, when the difference is large, the Gaussian kernel function approaches 0. This essentially means that when a potentially changed data point (for e.g., $\bold{q}_k$ ), is far varied from $\bold{p}_j$, the Gaussian kernel signifies that the points $\bold{p}_j$ and $\bold{q}_k$ are not similar. Intuitively, this similarity is dependent on the value of $\sigma$. This technique allows us to reject the faulty data point (shot noise or non-Gaussian noise) $\bold{q}_k$. This idea of calculating the maximum similarity as given in Equ. (\ref{Gsigma}) is called the maximum Correntropy criterion. 

When the value of $\bold{q}_k$ approaches $\infty$ then the kernel function approaches 0, i.e.,
\begin{equation}
   \lim_{\bold{q}_k \to \infty}G_\sigma=0.
    \label{kp1}
\end{equation} 
Intuitively, when the Gaussian Kernel shrinks to $0$, the Correntropy approaches the value $p(\bold{p}_j=\bold{q}_k)$~\cite{10.5555/1855180}.

\section{Proposed Method}
\label{Section_3}

\subsection{Correntropy Similarity Matrix with Iterative Closest Point Algorithm}
Given the general description of the Correntropy concept in the previous section, we extend this idea of Correntropy to the well known ICP algorithm. Traditional ICP ~\cite{10.1109/TPAMI.1987.4767965} describes the alignment of point clouds such that the Mean Square Error (MSE) between point sets is minimized. The MSE is the key criterion of convergence in ICP and its variants. ICP revolves around the problem of aligning points sets $\bold{{P}_s}$ and $\bold{{P}_t}$. We denote the point sets as $\bold{P}_{s}=\left \{[\bold{p}_j]_{j=1}^{N}\right \}$ (here $\left \{  \right \}$ denotes a point set) and $\bold{P}_{t}=\left \{[\bold{q}_k]_{k=1}^{N}\right \}$, respectively. Our goal is to best align \textit{Source} point set $\bold{{P}_s}$ to \textit{Target} point set (or a model point set) $\bold{{P}_t}$. This leads to finding the appropriate rotation and translation between the two point sets, which essentially means how much units the \textit{Source} dataset needs to be translated ($\bold{tr}$) and rotated ($\bold{R}$) such that \textit{Source} point set $\bold{{P}_s}$ is best aligned with $\bold{{P}_t}$ given that it satisfies a particular criterion (Root Mean Square Error (RMSE) threshold). This process is commonly known as registration between the \textit{Source} $\bold{{P}_s}$ and the \textit{Target} $\bold{{P}_t}$ is defined as finding the translation $\bold{tr}$ and the rotation $\bold{R}$ component such that $\bold{{P}_s}$ is in best alignment with $\bold{{P}_t}$. 

Now, the problem can be written as: find the best $\bold{R}$ and $\bold{tr}$ such that the MSE, $\varepsilon^{2}$, between 2 point clouds is minimized. This problem can be mathematically written as,
\begin{equation}
\varepsilon^{2}=\sum_{j,k=1}^{N}\left \| \bold{p}_j -(s\bold{R}\bold{q}_k+\bold{tr})\right \|^2,
\label{MSE1}
\end{equation} 
where $s$ is the scaling component. $\bold{p}_j \in \bold{{P}_{s}}$ is the set of all the points in the \textit{Source} point set, and $\bold{q}_k \in \bold{P}_{t}$ is the set of all the points in the \textit{Target} point set ($j,k=1,...,N$).

Our work is built on the work of Arun et al.\cite{10.1109/TPAMI.1987.4767965} and it is well known from their work that initially we compute the corresponding points between \textit{Source} and \textit{Target} point sets. The Corresponding points in this context are computed as shortest point from each point in the \textit{Source} dataset to the \textit{Target} dataset. This can be efficiently computed using k-d tree data structure. We introduce an additional parameter using the Correntropy criterion to find the similarity metric between points. Now, we say that for a \textit{Source} point index $i$ the corresponding point index in the \textit{Target} point set is  $\textbf{c}(i)$ ($\textbf{c}$ is vector of corresponding points from \textit{Source} to \textit{Target}. Note: We do not perform reciprocal correspondence i.e., from \textit{Target} to \textit{Source}). From Equ.\ref{Gsigma} we compute the similarity between the two corresponding points $\bold{{P}_s}$ and $\bold{{P}_t}$ as
\begin{equation}
\begin{split}
    \textbf{d}&=\bold{{P}_s}(i)-\bold{{P}_t}(\textbf{c}(i)), \\
    G_\sigma(\bold{{P}_s}(i)-\bold{{P}_t}(\textbf{c}(i)))&=\frac{1}{(2\pi\sigma)^{\frac{1}{D}}}exp(\frac{\textbf{d}^{T}\textbf{d}}{2\sigma^2}).
\end{split}
\label{sim1}
\end{equation}
Here, $D$ is the dimensions and $D=3$ in this case \cite{itlarticle}.Now, we can create a similarity matrix between the \textit{Source} and the \textit{Target} points based on the similarity metric computed in Equ. \ref{sim1}. Intuitively, the size of the similarity matrix is $N\times N$. In every iteration we initialize the value of this similarity matrix as zeros and update them in every iteration based on the similarity metric computed above in Equ.\ref{sim1}. It is updated as shown below:
\begin{equation}
\begin{split}
    \textbf{SM}(i,\textbf{c}(i))=G_\sigma(\bold{{P}_s}(i)-\bold{{P}_t}(\textbf{c}(i))), \\
    \textbf{SM}(\textbf{c}(i),i)=G_\sigma(\bold{{P}_s}(i)-\bold{{P}_t}(\textbf{c}(i))),
\end{split}
\label{sim2}
\end{equation}
where $\textbf{SM}$ is the similarity matrix which we intend to use in the computation of the rotation component. The centroid of both the point clouds i.e., the \textit{Source} and \textit{Target} is computed as given by Equ.\ref{CentS1}

\begin{equation}
\begin{split}
\bold{s_{cen}}&=\frac{1}{N}\sum_{i=1}^{N}\textbf{p}_i,  \\
\bold{t_{cen}}&=\frac{1}{N}\sum_{i=1}^{N}\textbf{q}_i,
 \end{split}
\label{CentS1}
\end{equation} 
where $\bold{s_{cen}}$ and $\bold{t_{cen}}$ are the computed centroid of \textit{Source} and  \textit{Target} data set respectively.
The difference between the computed centroid $\bold{s_{cen}}$ and the individual \textit{Source} points is computed. The same procedure is followed for the \textit{Target} points. It is given in Equ.~\ref{diff1}
\begin{equation}
\begin{split}
\textbf{p}_i^{'}&=\textbf{p}_i-\bold{s_{cen}}, \\
\textbf{q}_j^{'}&=\textbf{q}_j-\bold{t_{cen}}. 
 \end{split}
\label{diff1}
\end{equation} 
The Singular Value decomposition (SVD) for finding the rotation component is based on first finding the $4x4$ matrix as shown in Equ.~\ref{svd1} where we introduce our Similarity Matrix 
\begin{equation}
\begin{split}
\textbf{H}=\sum_{i=1}^{N}\textbf{p}_{i}^{'}\textbf{SM}\textbf{q}_{i}^{'T}.
 \end{split}
\label{svd1}
\end{equation} 
The size of the $\textbf{SM}$ matrix is $N \times N$ where as the size of the \textbf{p}' matrix is $4 \times N$ (under homogeneous transformation) which results in $4 \times N $ matrix. The result when multiplied with $\textbf{q}'^{T}$ (which is a $N \times 4$ matrix) will have the final result in a $4 \times 4$ matrix form which is the size of the Matrix $\textbf{H}$.
The rest of the algorithm is the same as the traditional algorithm where we compute the SVD of $\textbf{H}$ as given by Equ.~\ref{svd2}
\begin{equation}
\begin{split}
\textbf{H}=\textbf{U} \Lambda \textbf{V}^{T}.
 \end{split}
\label{svd2}
\end{equation}
Now the Rotation component is calculated as given by Equ.~\ref{svd3}.
\begin{equation}
\begin{split}
\textbf{R}=\textbf{V}\textbf{U}^{T}.
 \end{split}
\label{svd3}
\end{equation}
 The determinant of $R$ must be a positive integer. The translation component is simply computed as the difference of the centroids computed in Equ.\ref{diff1}. Algorithm \ref{CoSM_algo} describes the entire procedure of our proposed method.

\begin{algorithm}[!htb]
\begin{algorithmic}[1]
\Function{\textbf{ReadDataSets}}{$\bold{{P}_s}$,$\bold{{P}_t}$}\Comment{Read the \textit{Source} and the \textit{Target} datasets.}
 \While {not converged}
  \State $\textbf{c} \gets \textbf{ComputeCorrespondence} (\bold{P_s},\bold{P_t})$.\Comment{Compute Correspondence}
  \State  $\textbf{SM}=zeros(N,N)$\Comment{Initialize $N\times N$ Similarity Matrix to all zeros.}
   \For{$i \gets 1$ to $N$} 
      \State $\bold{d}=\bold{{P}_{s}}(i)-\bold{{P}_{t}}(\textbf{c}(i))$.
      \State Compute $G_\sigma$ as shown in Equ.\ref{sim1}.
      \State $\textbf{SM}(i,\textbf{c}(i))=G_\sigma(\bold{{P}_s}(i)-\bold{{P}_t}(\textbf{c}(i)))$.
      \State $\textbf{SM}(\textbf{c}(i),i)=G_\sigma(\bold{P_s}(i)-\bold{P_t}(\textbf{c}(i)))$.
   \EndFor
  \State Compute centroid as given in Equ.\ref{CentS1}.
  \State Compute the difference as given in Equ.\ref{diff1}.
  \State Compute the $\textbf{H}$ Matrix as given in Equ.\ref{svd1}.
  \State Compute SVD of $\textbf{H}$ as given in Equ.\ref{svd2}.
  \State Compute $\bold{R}$ as given in Equ.\ref{svd3}.
  \State Compute $\bold{tr}$ as difference of centroids.
 \EndWhile
\EndFunction
\end{algorithmic}
\caption{CoSM Algorithm}
\label{CoSM_algo}

\end{algorithm}

\subsection{Properties of the Similarity Matrix}
\begin{figure*}[t!]
 \centering
    \includegraphics[width=0.4\textwidth,height=4.3cm]{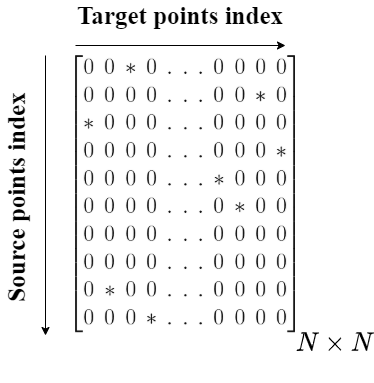} 
  \caption{Similarity Matrix.'*' represent the Correntropy values.}
  \label{fig:simmatrix}
\vspace{-0pt}
\end{figure*}

\subsubsection{$\textbf{SM}$ Matrix is a sparse Matrix}
Since at every iteration we initialize the values to zeros and based on Equ.~\ref{sim2} we fill the Correntropy similarity values at only specified location, the matrix is sparse. Line 7 and 8 in Algorithm \ref{CoSM_algo} clearly indicated that we update only those indices in the \textit{Source} and the \textit{Target} point indexes which are close to each other. Fig.~\ref{fig:simmatrix} shows a simple structure of the similarity matrix.
\subsubsection{The rows and columns are linearly independent}
\begin{figure*}[t!]
 \centering
    \includegraphics[width=0.4\textwidth,height=4cm]{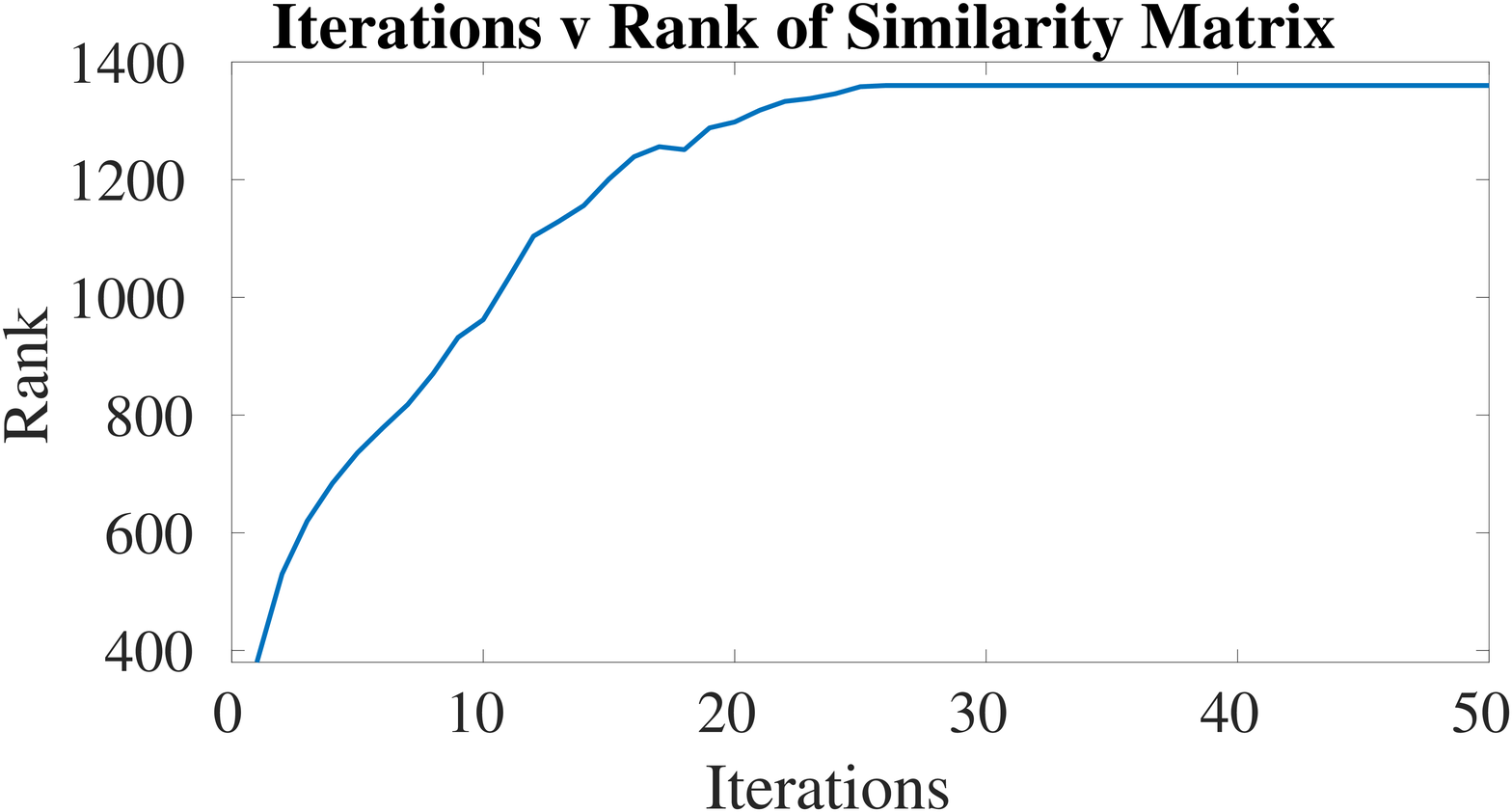} 
  \caption{  As the iterations increases, the rank of the matrix approaches $N$. Here $N$ is 1360.}
  \label{fig:rank1}
\vspace{-0pt}
\end{figure*}

Every element in the $\textbf{SM}$ matrix or $\textbf{SM}(i,j)$ represents the Correntropy relationship between the $i$th \textit{Source} point and the $j$th \textit{Target} point. Since we are computing the similarity of only the closest point from the \textit{Source} index to the \textit{Target} index, every other indices that particular row/column is intuitively zero. This clearly means for every row index in the \textit{Source} point set there is only one component in the \textit{Target} dataset with a similarity metric which assures one to one relationship between every \textit{Source} point to every \textit{Target} point. The points in the \textit{Source} point set are transformed based on this rotation matrix after every iteration. This process is repeated every iteration and one can find that the number of points with closer similarity metric $(\sim 1)$ increases. Fig~\ref{fig:rank1} shows that as the iterations increases the rank approaches $N$.

\subsubsection{Similarity Matrix is a Mirror-Symmetric Matrix. }
From Equ.~\ref{sim2}, one can verify that if the matrix is diagonally separated, then it's a mirror image of each other (Mirror-Symmetric Matrices) which means $\textbf{SM}=\textbf{SM}^{T}$.

\section{Results}\label{Results}
\subsection{Evaluation on Datasets with no outliers.}
We validated our approach on 3 simple and well known datasets: the Bunny Rabbit, the Happy Buddha and the Dragon dataset. In addition, we have also performed our evaluation on a Lidar dataset (KITTI dataset) since it captures data on a large scale. We perform our experiment on Intel i7 2.Ghz CPU with 32 Gb of RAM. We implemented our approach in the PCL library (version 1.9) and released our code in the github repository ( {\footurl}). In this section, we did not add any random outliers, and we introduce random rotation and translation between the \textit{Source} and the \textit{Target} point sets. First we capture a sample pcd file and transform the acquired point cloud with a random transformation matrix where all the individual components of the transformation matrix were randomly generated with different seeds (we assume throughout the paper that the units for angle are in radians). We name the original point cloud as read from the pcd file as \textit{Target} and the randomly transformed point cloud as the \textit{Source}. The problem statement is to find the relative transformation between the \textit{Source} and the \textit{Target} such that the \textit{Source} is aligned with the \textit{Target}. For the above steps, the algorithm is described in Algorithm ~\ref{random_transformation_algo}.

\begin{algorithm}[!htb]
\begin{algorithmic}[1]
\Function{\textbf{ReadPCDFile}}{'filename.pcd',$\bold{P_t}$}. \Comment{Read the PCD file and name it as a Target.}
    \State $a_{ax}$= Random angular component along the X-axis (-6.28 to 6.28 radians).
    \State $a_{ay}$= Random angular component along the Y-axis (-6.28 to 6.28 radians).
    \State $a_{az}$= Random angular component along the Z-axis (-6.28 to 6.28 radians).
    \State $tr_{tx}$= Random translation along the X-axis.
    \State $tr_{ty}$= Random translation along the Y-axis.
    \State $tr_{tz}$= Random translation along the Z-axis.
    \State $\textbf{M} \gets \textbf{GenerateTransformationMatrix}(a_{ax},a_{ay},a_{az},tr_{tx},tr_{ty},tr_{tz}$)
    \State $\bold{P_s} \gets \bold{TransformPointCloud}(\bold{P_t},\bold{M}$)
    \State \Return {$\bold{P_s}$} \Comment{Return the transformed point cloud and name it as Source.}

\EndFunction
\end{algorithmic}
\caption{Transform the point cloud with a Random Transformation Matrix.}
\label{random_transformation_algo}

\end{algorithm}

\begin{figure*}[!htb]
\centering
\subfloat[]{\includegraphics[width=0.3\textwidth,height=4cm]{Images/bun1.png}} \hfil
\subfloat[]{\includegraphics[width=0.3\textwidth,height=4cm]{Images/dr1.png}} \hfil
\subfloat[]{\includegraphics[width=0.3\textwidth,height=4cm]{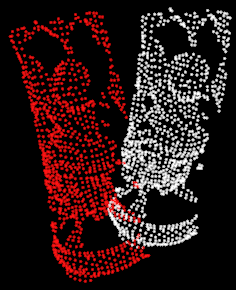}} \hfil
  \caption{Rotated and translated point cloud (white is the original point cloud (\textit{Target}), and Red is the transformed point cloud (\textit{Source})) with rotation of $(r,p,y)=(0.314,0,0)$  and translation of $0.05$ unit in the z-axis. (a) shows the bunny rabbit point cloud down sampled using voxel grid filtering of leaf size $0.005$. (b) shows the dragon point cloud down sampled using voxel grid filtering of leaf size $0.005$. (c) shows the happy buddha point cloud down sampled using voxel grid filtering of leaf size $0.005$.}
  \label{fig:Res1_samplepointclouds}
\vspace{-0pt}
\end{figure*}


\begin{figure*}[hb]
 \centering
    \subfloat[]{\includegraphics[width=0.3\textwidth,height=4cm]{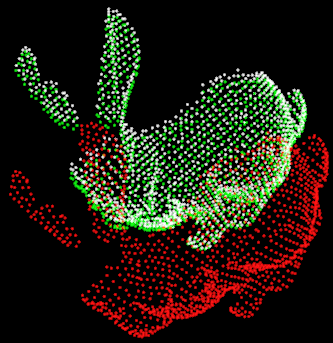}} \hfil
	\subfloat[]{\includegraphics[width=0.3\textwidth,height=4cm]{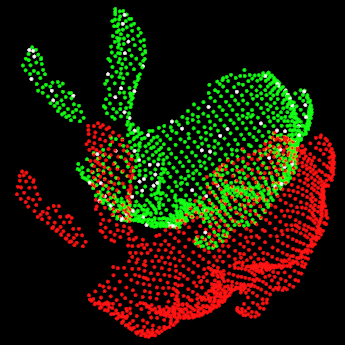}} \hfil
	\subfloat[]{\includegraphics[width=0.3\textwidth,height=4cm]{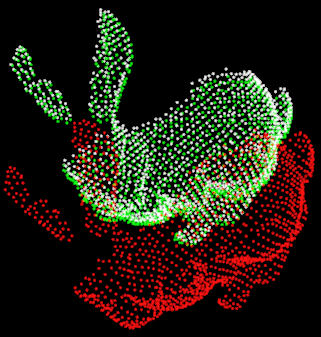}} \hfil
	\subfloat[]{\includegraphics[width=0.3\textwidth,height=4cm]{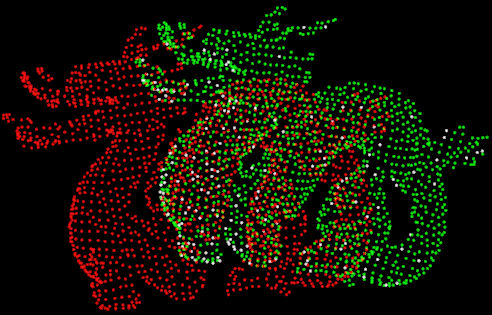}} \hfil
	\subfloat[]{\includegraphics[width=0.3\textwidth,height=4cm]{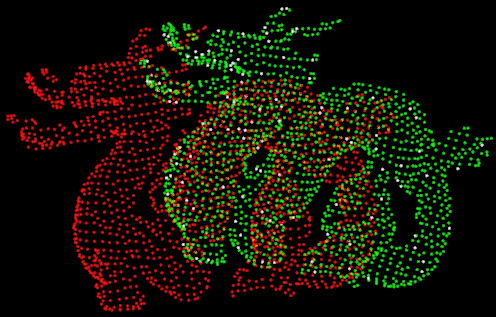}} \hfil
	\subfloat[]{\includegraphics[width=0.3\textwidth,height=4cm]{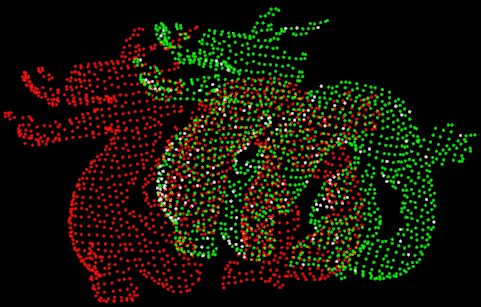}} \hfil
	\subfloat[]{\includegraphics[width=0.3\textwidth,height=4cm]{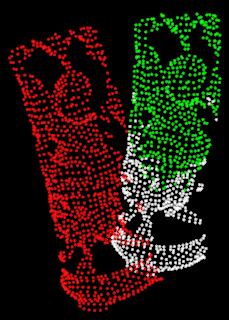}} \hfil
	\subfloat[]{\includegraphics[width=0.3\textwidth,height=4cm]{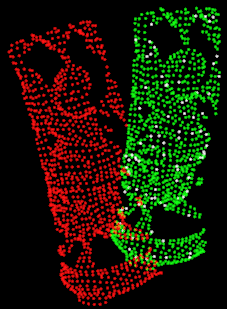}} \hfil
	\subfloat[]{\includegraphics[width=0.3\textwidth,height=4cm]{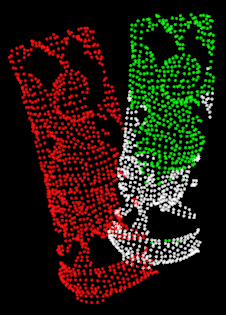}} \hfil
  \caption{Transformation from \textit{Source} to \textit{Target}: $(r,p,y,x,y,z)=(0.314,0,0,0,0,0.05)$. White point cloud: Original point cloud (\textit{Target}). Red point cloud: \textit{Source} point cloud and Green point cloud : \textit{Source} transformed point cloud after 10 iterations using (a) ICP Standard SVD (RMSE: 6.25494e-06) (b) ICP Point to Plane (RMSE: 2.77175e-16) and (c) CoSM ICP (RMSE: 3.34226e-06) on the Bunny Rabbit dataset. Similarly we perform 50 iterations on the Dragon dataset using (d) ICP Standard SVD (RMSE: 2.86089e-14 ) (e) ICP Point to Plane (RMSE: 3.93822e-16) and (f) CoSM ICP (RMSE: 2.72838e-15) and 35 iterations on the Happy Buddha dataset using (g) ICP Standard SVD (RMSE: 1.72851e-13) (h) ICP Point to Plane (RMSE: 3.91357e-16) and (i) CoSM ICP (RMSE: 2.46336e-14).}
  \label{fig:Res2_Buncomparison}
\vspace{-0pt}
\end{figure*}
\FloatBarrier

To validate our approach on the datasets, we collected the results on a simple scenario where the \textit{Source} is transformed manually with a simple transformation matrix which has less variation in rotation and translation ($(r,p,y,x,y,z)=(0.314,0,0,0,0,0.05)$). We reduce the number of point clouds to decrease the computational time by performing Voxel grid filtering (leaf size $0.005$) on the point clouds. The original number of points in the bunny point cloud sets is 40256 points, and after applying voxel grid filtering we get the number of points reduced to 1360 points. Similarly for the Dragon dataset, the original number of points is 43467 points, and after voxel grid filtering the number of points is reduced to 1593 points. The original number of points for the Happy Buddha dataset is 76166 points, and after voxel grid filtering we get 1090 points. The \textit{Source} and the \textit{Target} point clouds for the datasets are shown in Fig. \ref{fig:Res1_samplepointclouds}. 

\begin{figure*}[!htb]
 \centering
    \subfloat[]{\includegraphics[width=0.33\textwidth,height=3.3cm]{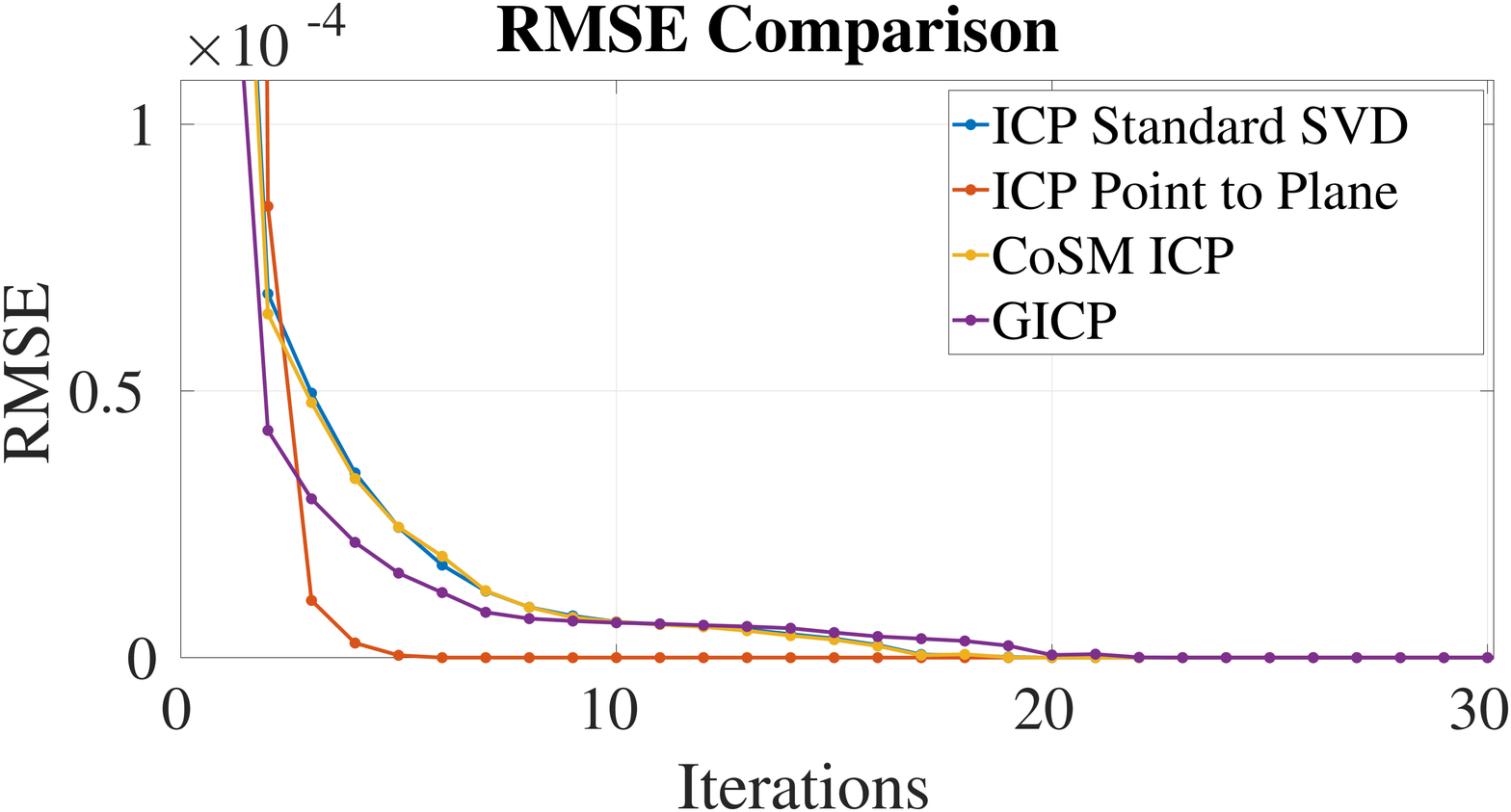}} \hfil
	\subfloat[]{\includegraphics[width=0.33\textwidth,height=3.3cm]{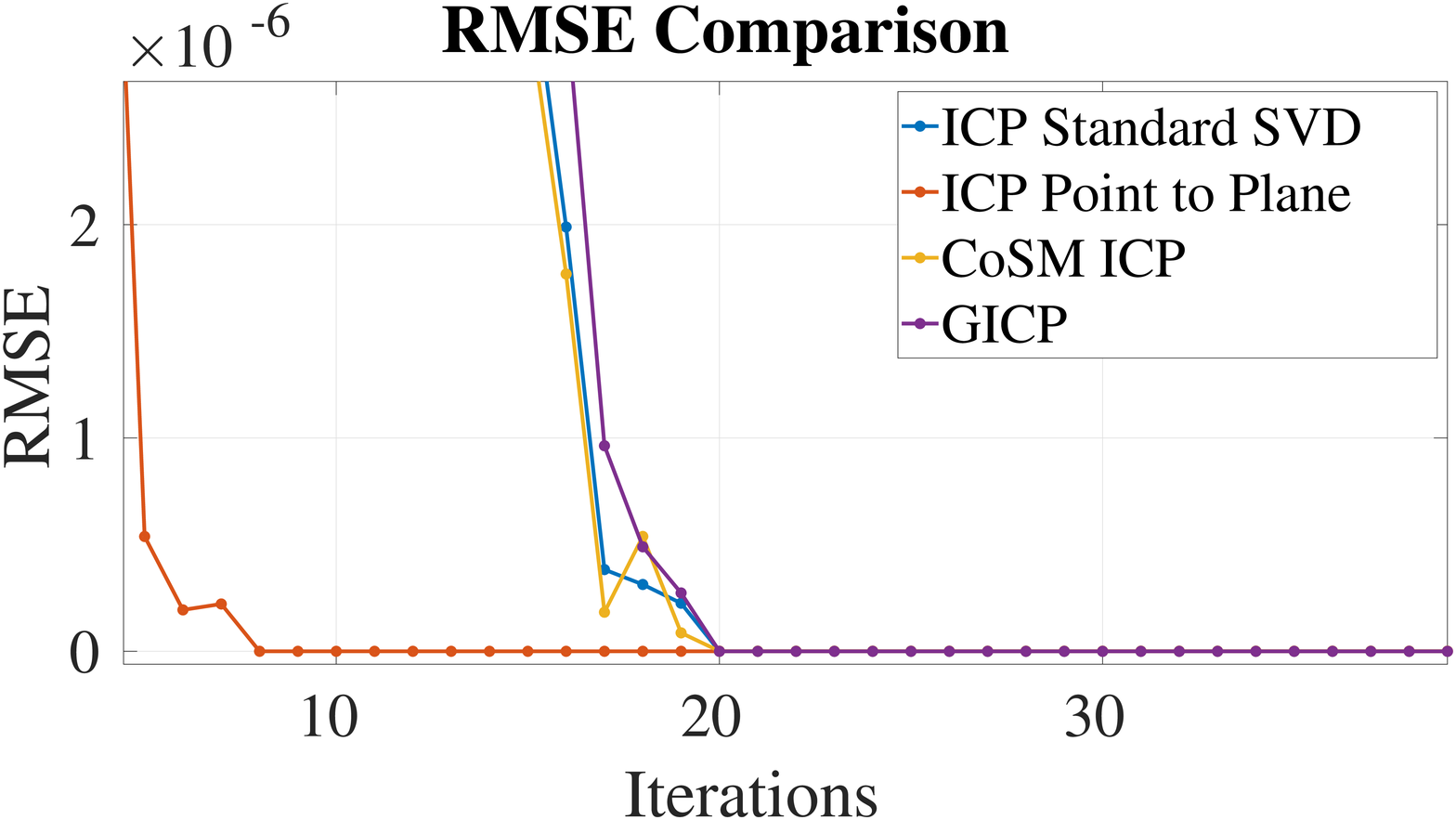}} \hfil
	\subfloat[]{\includegraphics[width=0.33\textwidth,height=3.3cm]{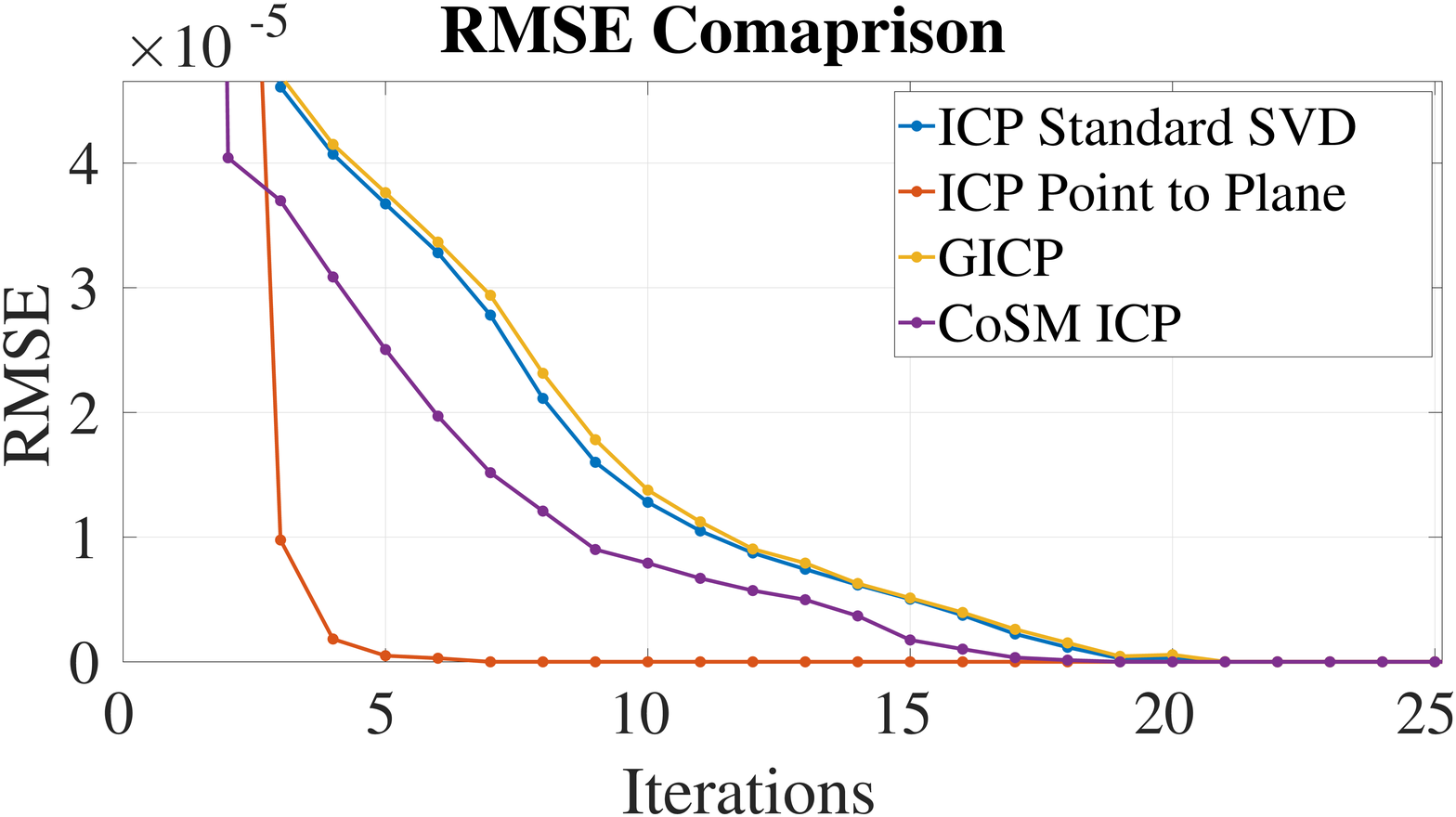}} \hfil
  \caption{Convergence of different Registration Methods (ICP Standard SVD, ICP point to plane, GICP and CoSM ICP) on different datasets on the simple transformation ($(r,p,y,x,y,z)=(pi/10,0,0,0,0,0.05)$). (a) Shows RMSE comparison on Bunny Rabbit dataset. (b) Shows RMSE comparison on the Dragon dataset. (c) RMSE comparison on the Happy Buddha dataset.}
  \label{fig:Res3_rmsecomp1}
\vspace{-0pt}
\end{figure*}

After applying few iterations, our approach aligns equally well (RMSE: 2.46336e-14) in comparison to the well known methods like ICP with standard SVD method (RMSE: 6.25494e-06) and ICP point to plane (RMSE: 2.77175e-16). The results for different datasets are shown in Fig.~\ref{fig:Res2_Buncomparison}. We compared the Root Mean Square Error (RMSE) on this particular transformation and we see that ICP point to plane converges faster than other methods including ours (in 50 iterations). Fig.~\ref{fig:Res3_rmsecomp1} shows the RMSE comparisons of different approaches on different datasets for this simple transformation. In this example, ICP point to plane seems to converge faster than our method since the rotation and the translation component is very small. The next step is to evaluate how well our method performs as compared to the other methods when the rotation is large.

\begin{figure*}[!htb]
 \centering
    \subfloat[]{\includegraphics[width=0.3\textwidth,height=4cm]{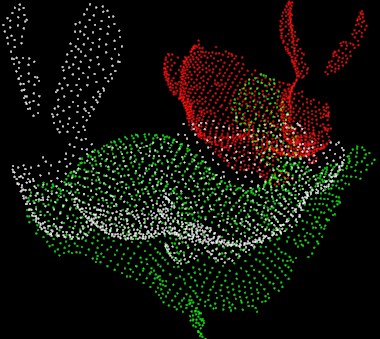}} \hfil
	\subfloat[]{\includegraphics[width=0.3\textwidth,height=4cm]{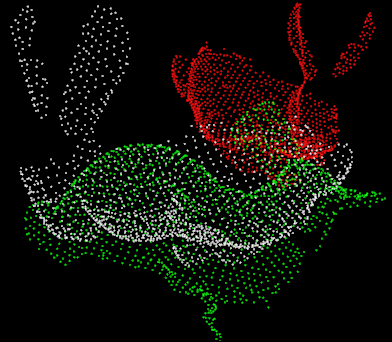}} \hfil
	\subfloat[]{\includegraphics[width=0.3\textwidth,height=4cm]{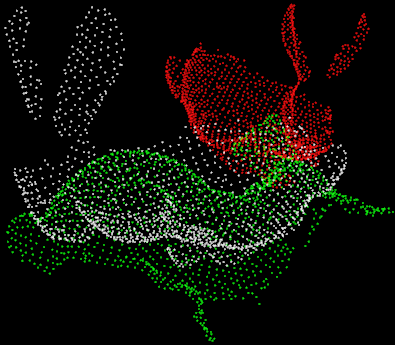}} \hfil
	\subfloat[]{\includegraphics[width=0.3\textwidth,height=4cm]{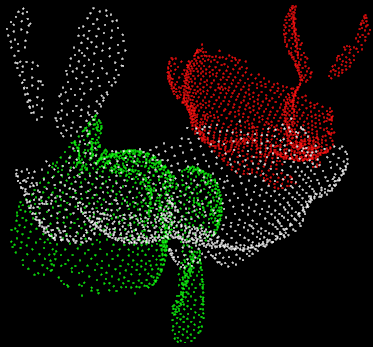}} \hfil
	\subfloat[]{\includegraphics[width=0.3\textwidth,height=4cm]{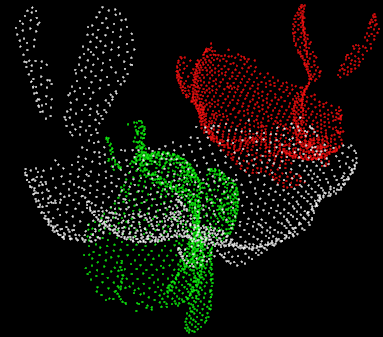}} \hfil
	\subfloat[]{\includegraphics[width=0.3\textwidth,height=4cm]{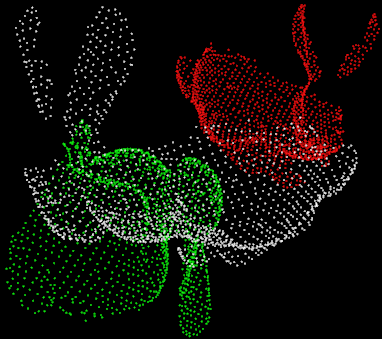}} \hfil
	\subfloat[]{\includegraphics[width=0.3\textwidth,height=4cm]{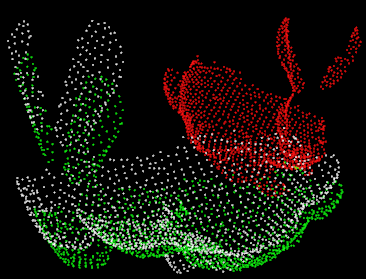}} \hfil
	\subfloat[]{\includegraphics[width=0.3\textwidth,height=4cm]{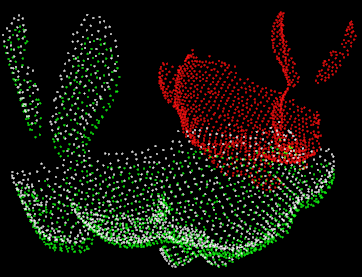}} \hfil
	\subfloat[]{\includegraphics[width=0.3\textwidth,height=4cm]{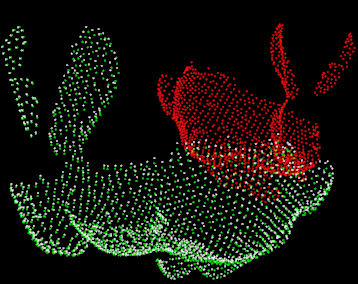}} \hfil
      \caption{White Point Cloud: Original Point Cloud (\textit{Target}). Red Point Cloud: \textit{Source} point cloud. Green Point Cloud: \textit{Source} transformed point cloud after applying iterations. \textit{Source} is transformed from the \textit{Target} as ($(r,p,y,x,y,z)=(-1.32811,-5.87854,2.12814,-0.874,-0.433,0.221)$). (a)-(c) shows the convergence of the \textit{Source} point clouds after 5, 10 and 20 iterations using the standard ICP,  respectively (RMSE after 20 iterations is 0.000358474). (d)-(f) shows the same using ICP Point to Plane (RMSE after 20 iterations is 0.00216325). (g)-(i) shows the same using CoSM ICP (RMSE after 20 iterations is 4.76074e-06).  }
  \label{fig:Res4_largecomparison}
\vspace{-0pt}
\end{figure*}

Fig.~\ref{fig:Res4_largecomparison} shows the results of different approaches including ours when the rotation and the translation component of the transformation matrix in quite large. In this example we use the Bunny Rabbit dataset where the \textit{Source} is transformed from the \textit{Target} as:

$(r,p,y,x,y,z)=(-1.32811,-5.87854,2.12814,-0.874,-0.433,0.221)$. 

\begin{figure*}[!htb]
 \centering
    \subfloat[]{\includegraphics[width=0.48\textwidth,height=5cm]{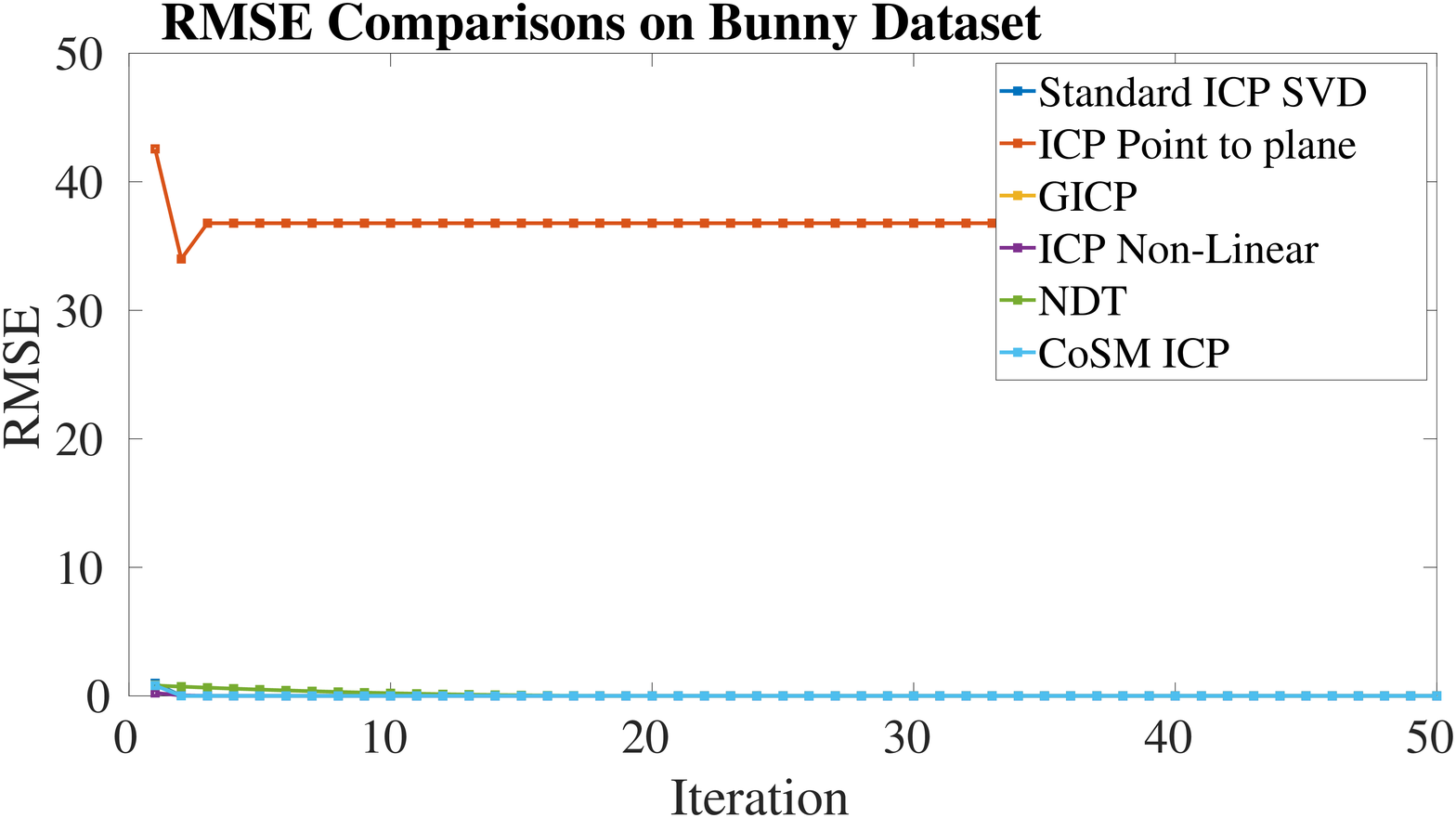}} \hfil
	\subfloat[]{\includegraphics[width=0.48\textwidth,height=5cm]{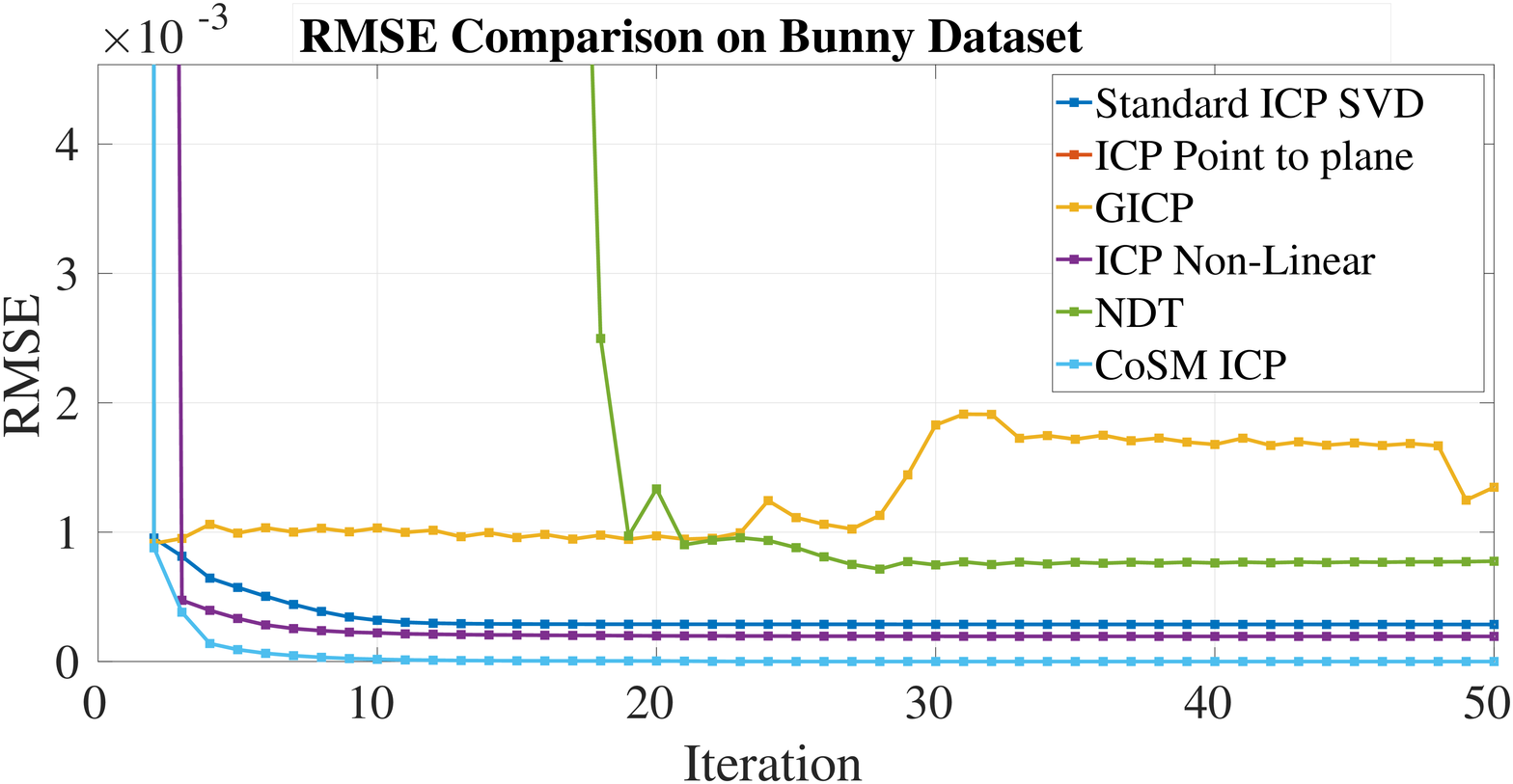}} \hfil
	\caption{RMSE comaprison of various methods on the Bunny Rabbit dataset. (b) shows the zoomed in version of (a). }
  \label{fig:Res5_rmsecompbun1}
\vspace{-0pt}
\end{figure*}

Clearly the translation component is small as compared to the rotation component. In this example, our method performs well in comparison to the other methods with RMSE as 4.76074e-06 in our method after 20 iterations. It clearly shows that the \textit{Source} is aligned very well with our approach where as in other methods the \textit{Source} failed to align with the \textit{Target} point cloud. We compared the RSME results with other approaches like GICP, NDT and ICP-nonlinear \cite{inproceedingsgicp,ndt1249285,FITZGIBBON20031145}. Fig~\ref{fig:Res5_rmsecompbun1} shows the RMSE results for various methods on the Bunny Rabbit dataset. It is clear that our method performs well in comparison to others. In addition, Table~\ref{Tab:RMSE_Compare1} shows the RMSE comparison of various methods applied on the Bunny Rabbit dataset with various rotation and translation.

\begin{figure*}[!htb]
 \centering
    \subfloat[]{\includegraphics[width=0.3\textwidth,height=4cm]{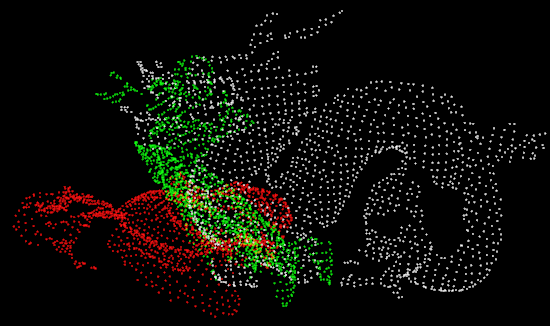}} \hfil
	\subfloat[]{\includegraphics[width=0.3\textwidth,height=4cm]{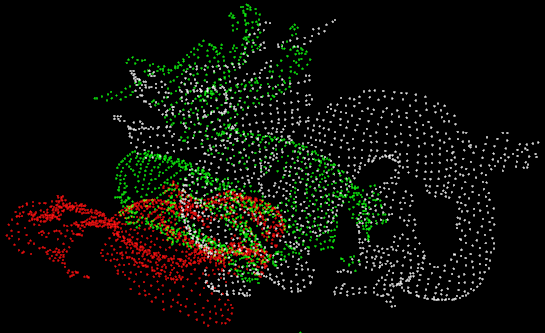}} \hfil
	\subfloat[]{\includegraphics[width=0.3\textwidth,height=4cm]{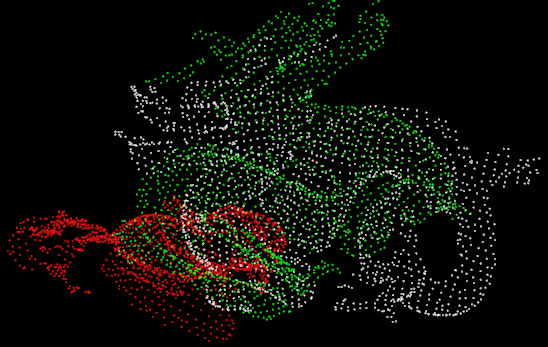}} \hfil
	\subfloat[]{\includegraphics[width=0.3\textwidth,height=4cm]{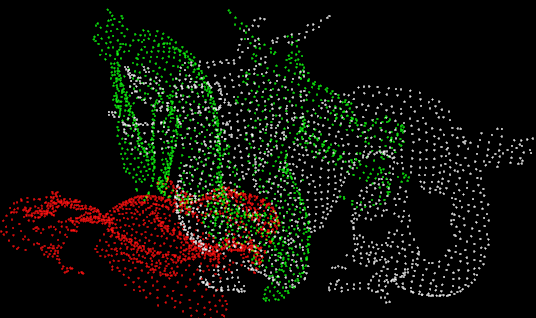}} \hfil
	\subfloat[]{\includegraphics[width=0.3\textwidth,height=4cm]{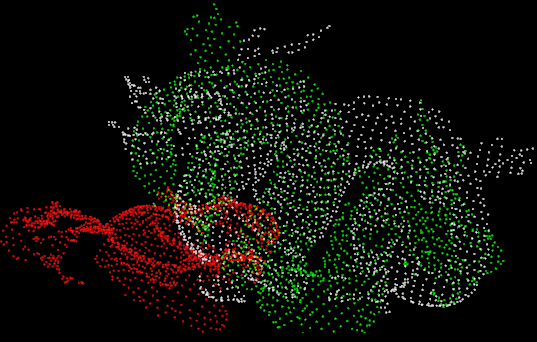}} \hfil
	\subfloat[]{\includegraphics[width=0.3\textwidth,height=4cm]{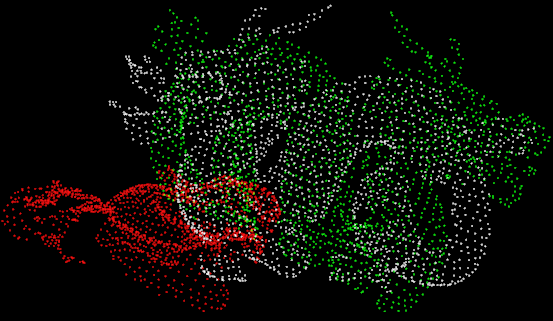}} \hfil
	\subfloat[]{\includegraphics[width=0.3\textwidth,height=4cm]{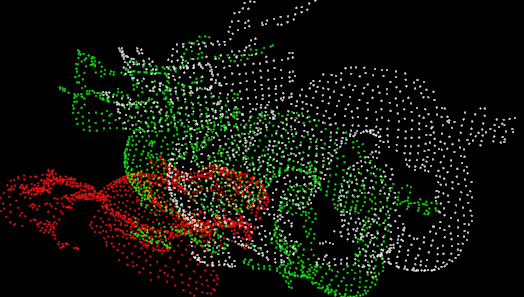}} \hfil
	\subfloat[]{\includegraphics[width=0.3\textwidth,height=4cm]{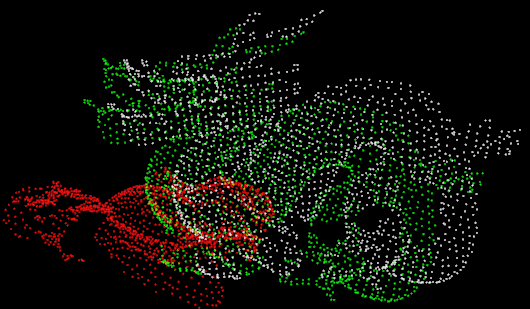}} \hfil
	\subfloat[]{\includegraphics[width=0.3\textwidth,height=4cm]{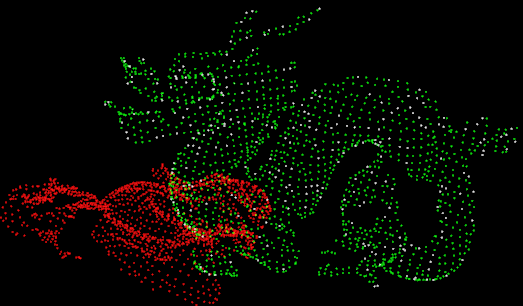}} \hfil
      \caption{White point cloud: Original Point Cloud (\textit{Target}). Red point cloud: \textit{Source} point cloud. Green point cloud: shows the \textit{Source} transformed point cloud after applying different registration methods. \textit{Source} is transformed from the \textit{Target} as ($(r,p,y,x,y,z)=(2.39318,-5.02554,-2.69076,0.000,-0.003,0.003)$). (a)-(c) shows the convergence of the \textit{Source} point clouds after 5, 10 and 25 iterations using the standard ICP, respectively (RMSE after 25 iterations is 0.000198014). (d)-(f) shows the same using ICP Point to Plane (RMSE after 25 iterations is 0.000136451). (g)-(i) shows the same using CoSM ICP (RMSE after 25 iterations is 8.66664e-08).  }
  \label{fig:Res5_largecomaprisondr2}
\vspace{-0pt}
\end{figure*}

In Fig.~\ref{fig:Res5_largecomaprisondr2}  we use the Dragon dataset where the \textit{Source} is transformed from the \textit{Target} as:
  
  $(r,p,y,x,y,z)=(2.39318,-5.02554,-2.69076,0.000,-0.003,0.003)$. 
  
\begin{figure*}[!htb]
 \centering
    \subfloat[]{\includegraphics[width=0.48\textwidth,height=5cm]{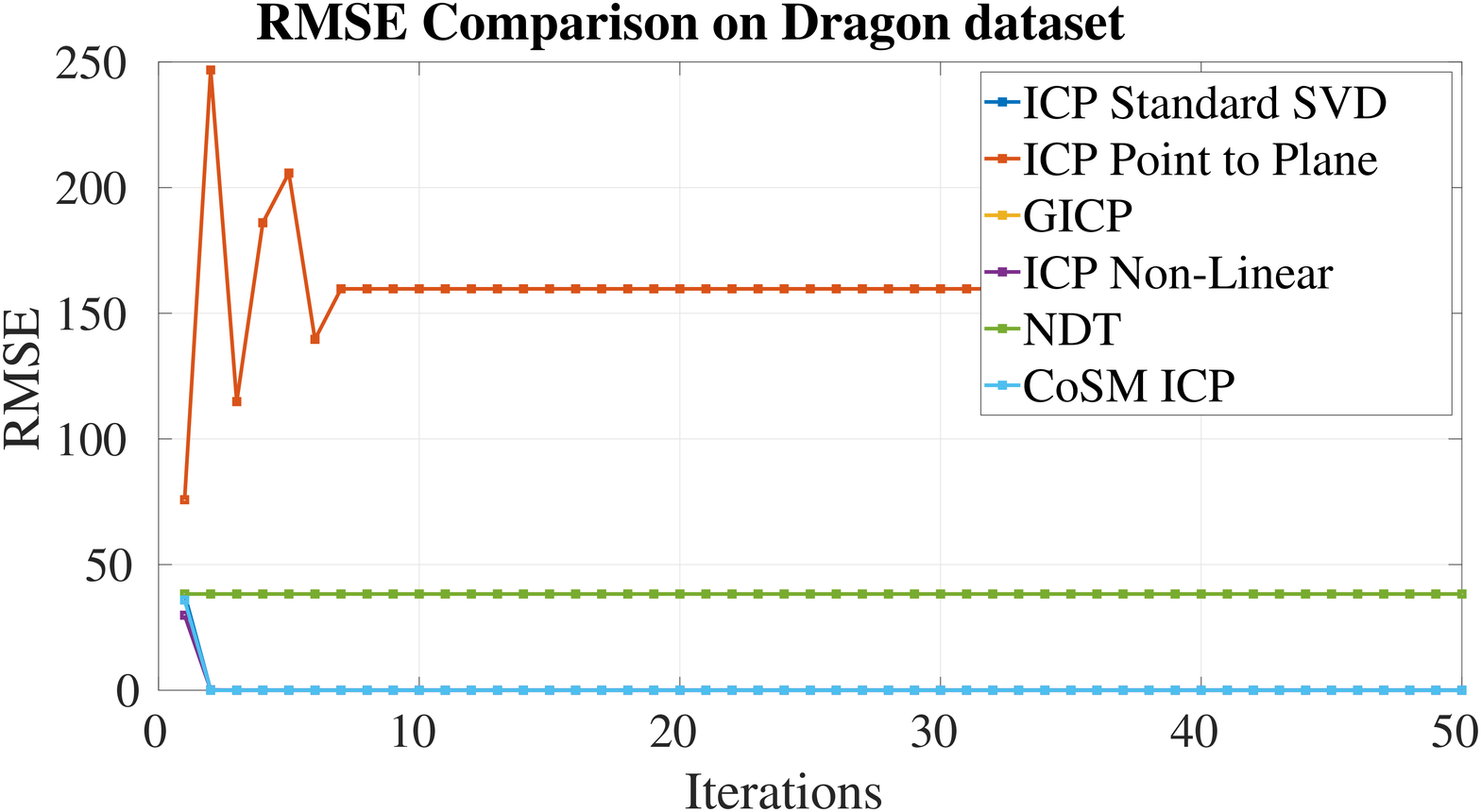}} \hfil
	\subfloat[]{\includegraphics[width=0.48\textwidth,height=5cm]{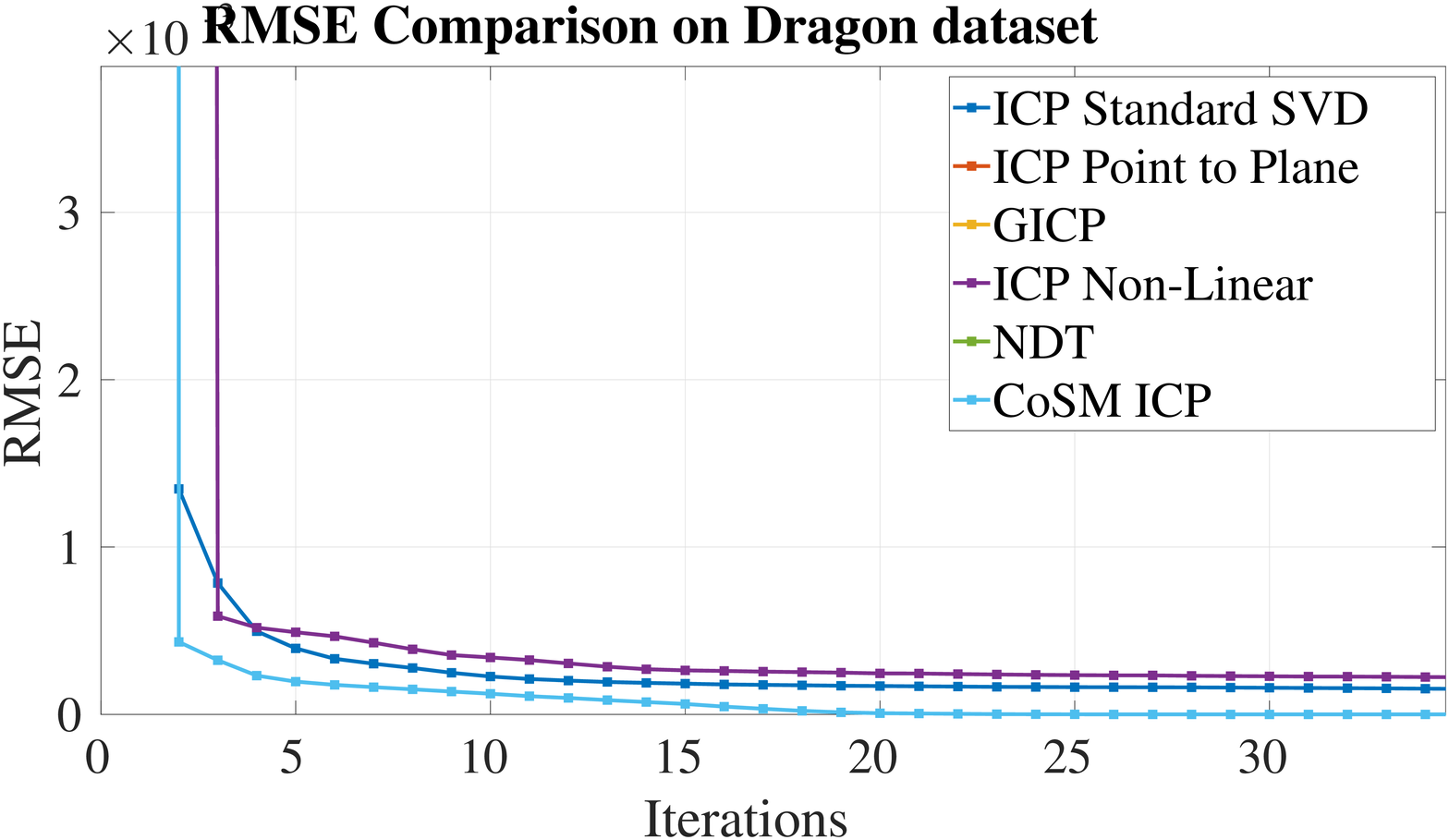}} \hfil
	
  \caption{RMSE comparison of various methods on Dragon dataset. (b) shows the zoomed in version of (a). }
  \label{fig:Res5_rmsecompdr1}
\vspace{-0pt}
\end{figure*}

  Again, our method performs well in comparison to other methods with RMSE as 8.66664e-08 in our method after 25 iterations. We arrive at a similar conclusion that the \textit{Source} is aligned very well with our approach where as in other methods the \textit{Source} failed to align with the \textit{Target} point cloud. Fig.~\ref{fig:Res5_rmsecompdr1} shows the RMSE results for various methods on the Dragon dataset. In addition, Table  \ref{Tab:RMSE_Compare2} shows the RMSE comparison of various methods applied on the Dragon dataset with various rotation and translation.

\begin{figure*}[!htb]
 \centering
    \subfloat[]{\includegraphics[width=0.3\textwidth,height=4cm]{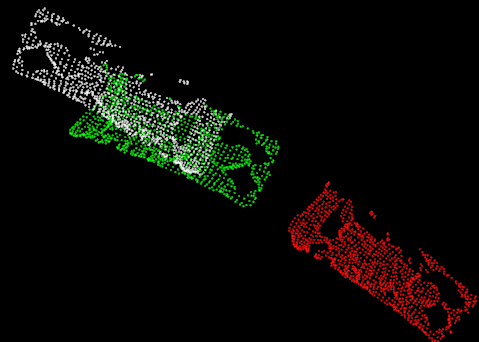}} \hfil
	\subfloat[]{\includegraphics[width=0.3\textwidth,height=4cm]{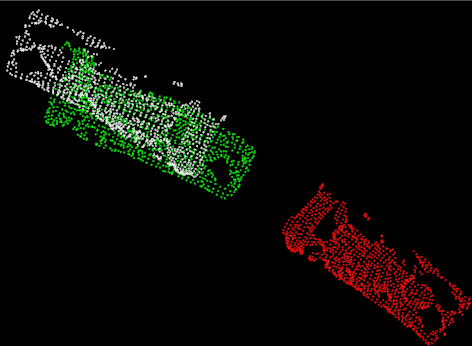}} \hfil
	\subfloat[]{\includegraphics[width=0.3\textwidth,height=4cm]{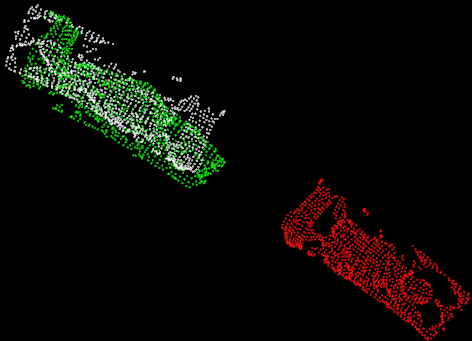}} \hfil
	\subfloat[]{\includegraphics[width=0.3\textwidth,height=4cm]{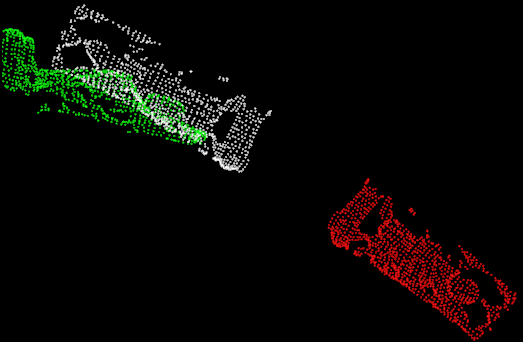}} \hfil
	\subfloat[]{\includegraphics[width=0.3\textwidth,height=4cm]{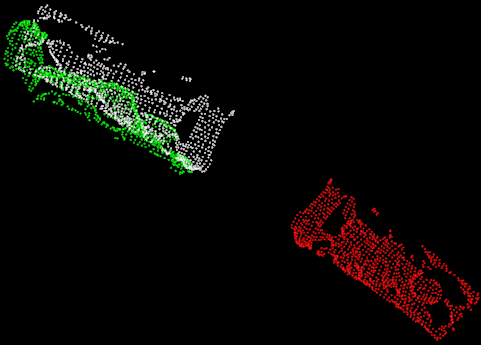}} \hfil
	\subfloat[]{\includegraphics[width=0.3\textwidth,height=4cm]{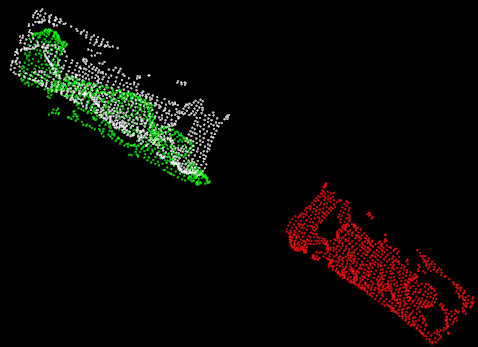}} \hfil
	\subfloat[]{\includegraphics[width=0.3\textwidth,height=4cm]{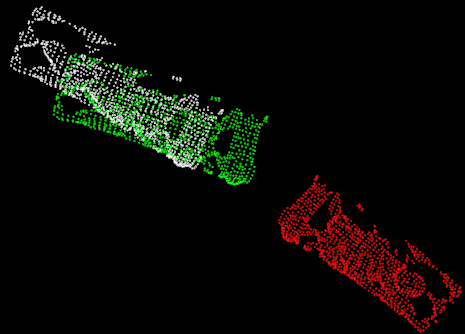}} \hfil
	\subfloat[]{\includegraphics[width=0.3\textwidth,height=4cm]{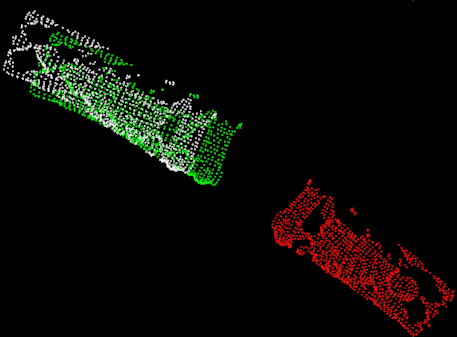}} \hfil
	\subfloat[]{\includegraphics[width=0.3\textwidth,height=4cm]{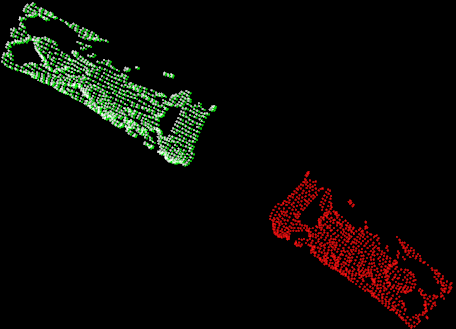}} \hfil
      \caption{White point cloud: Original point cloud (\textit{Target}). Red point cloud: \textit{Source} point cloud. Green: \textit{Source} transformed point cloud after applying different registration methods. Source is transformed from the \textit{Target} as ($(r,p,y,x,y,z)=(-4.50504,1.31677,4.83251,-0.023,-0.019,-0.008)$). (a)-(c) shows the convergence of the \textit{Source} point clouds after 5, 10 and 25 iterations using the standard ICP,  respectively (RMSE after 25 iterations is 0.000128905). (d)-(f) shows the same using ICP Point to Plane (RMSE after 25 iterations is 0.000124717). (g)-(i) shows the same using CoSM ICP (RMSE after 25 iterations is 2.26675e-06).  }
  \label{fig:Res6_hpdataset1}
\vspace{-0pt}
\end{figure*}

\begin{figure*}[!htb]
 \centering
    \subfloat[]{\includegraphics[width=0.48\textwidth,height=5cm]{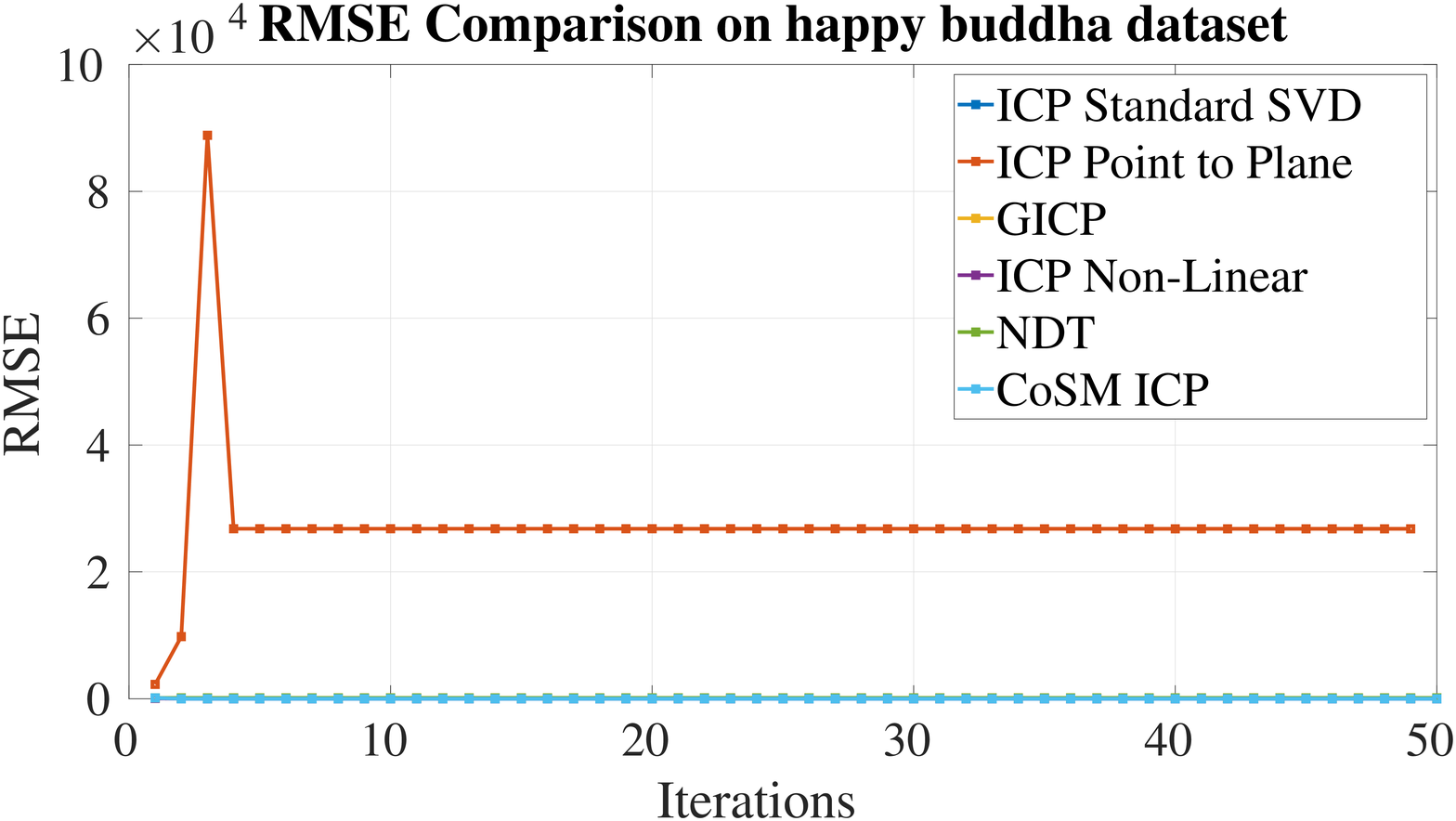}} \hfil
	\subfloat[]{\includegraphics[width=0.48\textwidth,height=5cm]{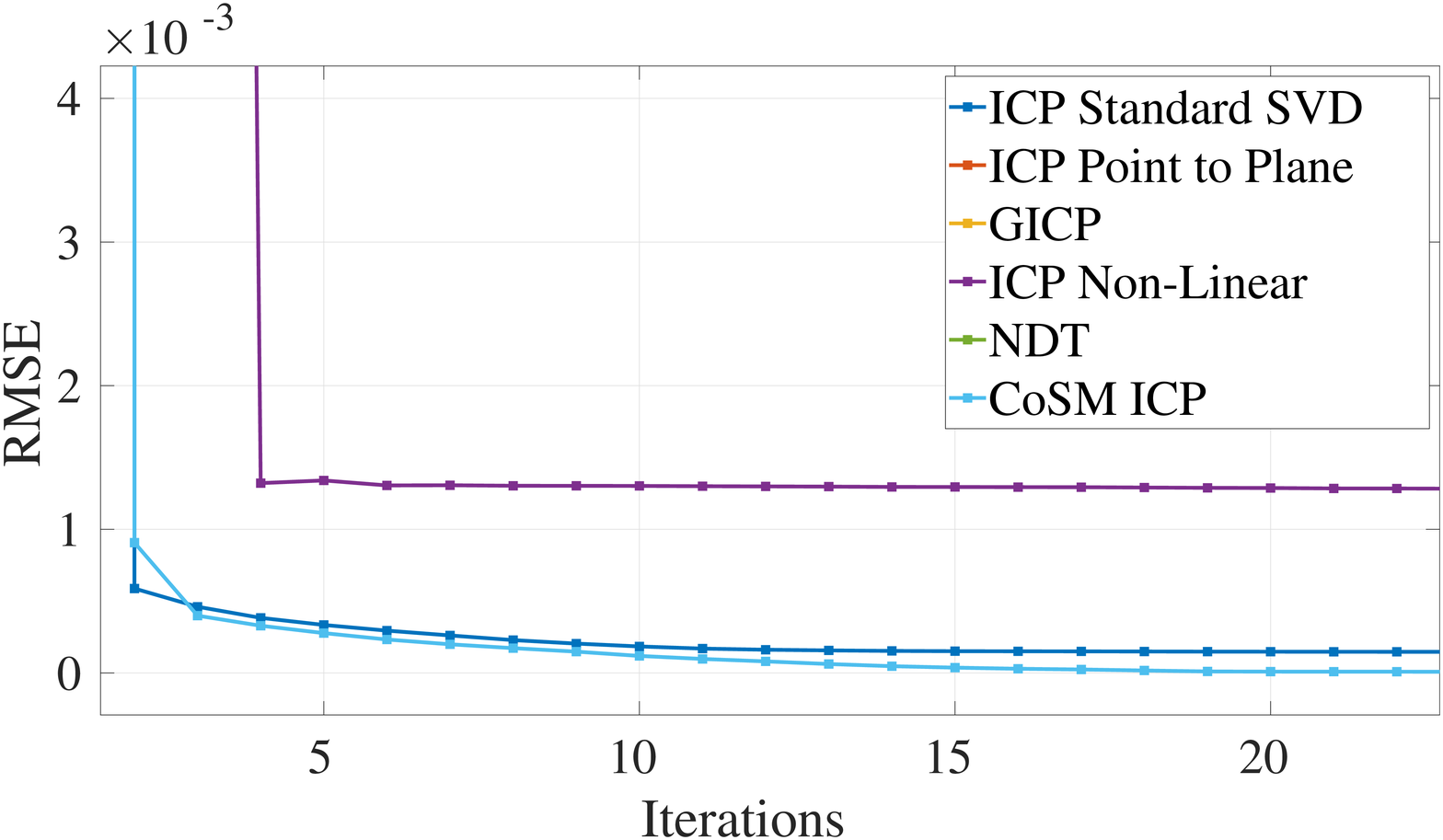}} \hfil
	
  \caption{RMSE comparison of various methods on the Happy Buddha dataset. (b) shows the zoomed in version of (a). }
  \label{fig:Res6_rmsecomphp1}
\vspace{-0pt}
\end{figure*}

In Fig.~\ref{fig:Res6_hpdataset1}  we use the Happy Buddha dataset where the \textit{Source} is transformed from the \textit{Target} as  $(r,p,y,x,y,z)=(-4.50504,1.31677,4.83251,-0.023,-0.019,-0.008)$.  Again, our method performs well in comparison to the rest of the methods with RMSE as 2.26675e-06 in our method after 25 iterations. We arrive at a similar conclusion that the \textit{Source} is aligned very well with our approach where as in other methods the \textit{Source} failed to align with the \textit{Target} point cloud. Fig~\ref{fig:Res6_rmsecomphp1} shows the RMSE results for various methods on  Dragon dataset. In addition, Table  \ref{Tab:RMSE_Compare3} shows the RMSE comparison of various methods applied on Dragon dataset with various rotation and translation.

\begin{table}

\caption{RMSE comparison of different methods with different values of rotation and translation after 50 iterations on Bunny Rabbit dataset. Note: Rotation component is in radians.}
\scalebox{0.7}[0.7]{%
\begin{tabular}{ |c|c|c|c|c|c|c| } 
\hline
  Transformation(r,p,y,x,y,z) & ICP Standard SVD & ICP Point to Plane & GICP & ICP Non-Linear & NDT & CoSM ICP\\
\hline
(2.39384,-2.57132,4.66973,0.876204,-2.83931,2.68268) & 0.000428388 & 88.0656 & 0.0011591 & 0.000206335 & 15.5141 & 1.81066e-13\\
(6.10518,-0.249119,2.41527,1.99458,8.99637,1.20097) & 0.000274292 & 268.684 & 81.2352 & 0.000301596 & 81.2352 & 3.09602e-13\\
(1.17438,-5.95203,-4.13622,4.6532,6.28659,0.0542642) & 0.000177495 & 147.987 & 58.9048 & 2.69103e-11 & 58.9048 & 1.75634e-13\\
(-0.866749,-2.6182,-0.318386,-2.1561,-1.25001,-4.8753) & 0.000283007 & 859.114 & 28.7628 & 0.000244988 & 28.7628 & 1.98801e-13\\
(5.08434,-3.9644,-2.66895,2.45251,-6.82633,1.41512) & 0.000471378 & 58.1505 & 55.996 & 1.54e-11 & 55.996 & 2.63875e-13\\
\hline
\end{tabular}}
\label{Tab:RMSE_Compare1}
\end{table}

\FloatBarrier

\FloatBarrier

\begin{table}

\caption{RMSE comparison of different methods with different values of rotation and translation (from \textit{Source} to \textit{Target}) after 50 iterations on Dragon dataset. Note: Rotation component is in radians}
\scalebox{0.7}[0.7]{%
\begin{tabular}{ |c|c|c|c|c|c|c| } 
\hline
  Transformation(r,p,y,x,y,z) & ICP Standard SVD & ICP Point to Plane & GICP & ICP Non-Linear & NDT & CoSM ICP\\
\hline
(-2.8152,-2.09474,-2.48844,5.66472,6.31125,-9.44663) & 0.000167091 & 638.278 & 156.91 & 0.000159398 & 156.9 & 4.19748e-13\\
(0.429847,2.60943,3.61045,3.6405,-1.63246,7.60966) & 0.00016037 & 325.136 & 73.1995 & 0.000224256 & 73.1995 & 1.25329e-13\\
(-2.78018,2.15172,-4.43629,-4.52232,-9.42573,-2.54953) & 0.000142098 & 3015.76 & 115.142 & 0.000162551 & 115.142 & 3.78298e-13\\
(1.75465,-1.56088,-1.70889,2.33321,6.13018,3.79368) & 0.000142188 & 26.6913 & 56.4895 & 0.000224482 & 56.4895 & 1.94386e-13\\
(-3.72204,1.05945,2.34622,-4.65481,9.74086,3.64636) & 0.000159037 & 115.65 & 127.159 & 126.91 & 127.159 & 4.19583e-13\\
\hline
\end{tabular}}
\label{Tab:RMSE_Compare2}
\end{table}

\begin{table}

\caption{RMSE comparison of different methods with different values of rotation and translation (from \textit{Source} to \textit{Target})after 50 iterations on the Happy Buddha dataset. Note: Rotation component is in radians}
\scalebox{0.7}[0.7]{%
\begin{tabular}{ |c|c|c|c|c|c|c| } 
\hline
  Transformation(r,p,y,x,y,z) & ICP Standard SVD & ICP Point to Plane & GICP & ICP Non-Linear & NDT & CoSM ICP\\
\hline
(0.429847,2.60943,3.61045,3.6405,-1.63246,7.60966) & 0.00016037 & 325.136 & 73.1995 & 0.000224256 & 73.1995 & 1.25329e-13\\
(-1.58692,-5.66055,-4.80461,-7.03304,0.718292,2.77415) & 0.000142378 & 469.843 & 55.5392 & 0.00016041 & 55.5392 & 1.41924e-13\\
(-2.79112,4.31946,-2.15715,2.44105,2.03992,-1.87023) & 0.000121835 & 39.1843 & 0.0018519 & 0.000133238 & 11.981 &  7.99904e-14\\
(-0.744191,-1.86187,-1.78262,-4.61717,1.79701,-7.04776) & 0.000192252 & 81.0697 & 71.9632 & 0.000192185 & 71.9632 & 1.80283e-13\\
(5.17263,1.57234,1.9238,-0.263428,5.05962,-2.27954) & 0.000157366 & 116.57 & 27.6866 & 0.000182242 & 27.6866 & 1.38844e-13\\
\hline
\end{tabular}}
\label{Tab:RMSE_Compare3}
\end{table}

\begin{figure*}[!htb]
 \centering
\includegraphics[width=0.7\textwidth,height=5cm]{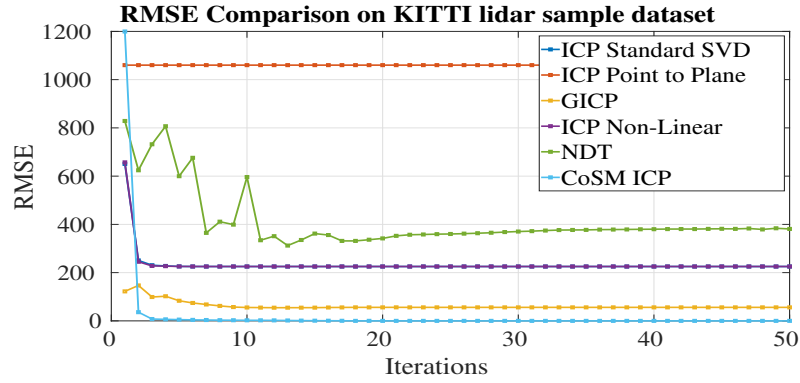}
	
  \caption{RMSE comparison of various methods on the KITTI Lidar dataset. }
  \label{fig:Res7_rmsecompli1}
\vspace{-0pt}
\end{figure*}

 For evaluating on the Lidar dataset we applied a voxel grid filtering of leaf size $1.5$. The original point cloud size is 115150 points and after voxel grid filtering we get 1497 points. In Fig.~\ref{fig:Res6_hpdataset1}  we use the Lidar dataset where the \textit{Source} is transformed from the \textit{Target} as  $(r,p,y,x,y,z)=(-4.27108,-0.505914,0.0988647,10.938, -10.532, 17.832)$.  Again, our method performs well in comparison to the rest of the methods with RMSE as 2.26675e-06 in our method after 25 iterations. We arrive at a similar conclusion that the \textit{Source} is aligned very well with our approach where as in other methods the \textit{Source} failed to align with the \textit{Target} point cloud. Fig~\ref{fig:Res7_rmsecompli1} shows the RMSE results for various methods on the Lidar dataset. In addition, Table  \ref{Tab:RMSE_Compare4} shows the RMSE comparison of various methods applied on the Lidar dataset with various rotation and translation.

\begin{figure*}[!htb]
 \centering
    \subfloat[]{\includegraphics[width=0.3\textwidth,height=4cm]{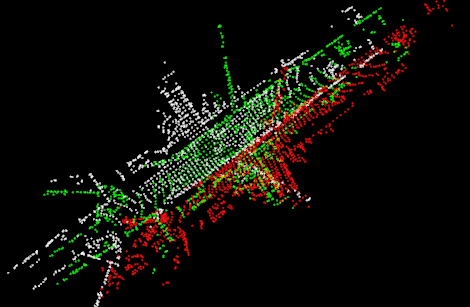}} \hfil
	\subfloat[]{\includegraphics[width=0.3\textwidth,height=4cm]{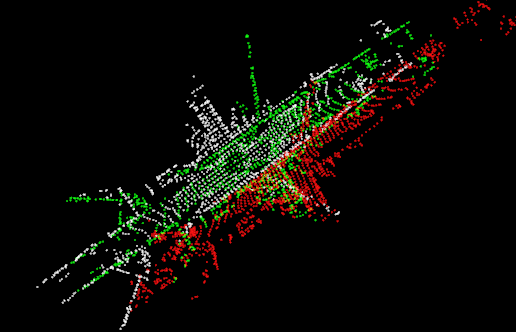}} \hfil
	\subfloat[]{\includegraphics[width=0.3\textwidth,height=4cm]{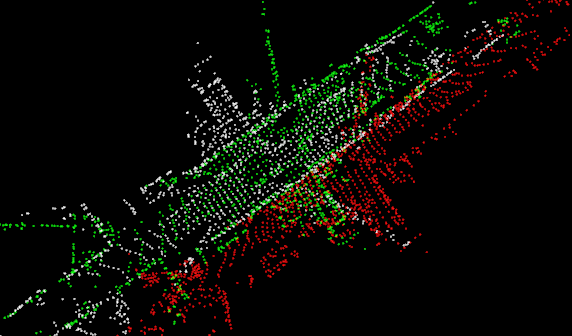}} \hfil
	\subfloat[]{\includegraphics[width=0.3\textwidth,height=4cm]{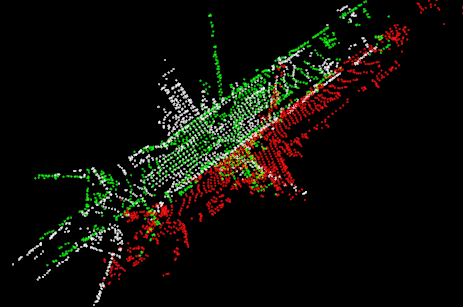}} \hfil
	\subfloat[]{\includegraphics[width=0.3\textwidth,height=4cm]{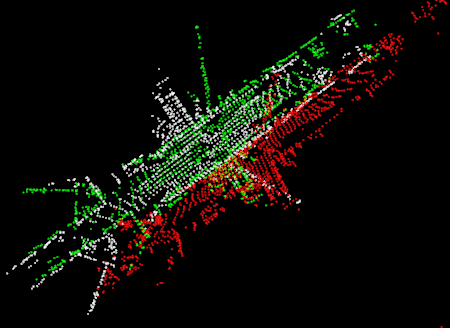}} \hfil
	\subfloat[]{\includegraphics[width=0.3\textwidth,height=4cm]{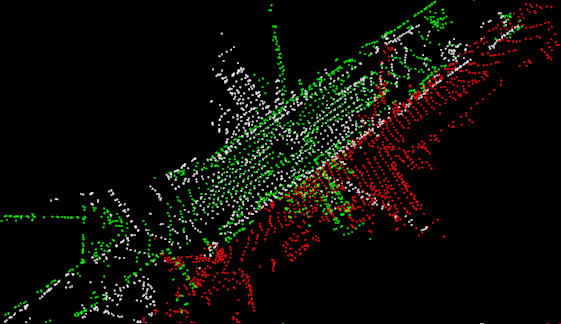}} \hfil
	\subfloat[]{\includegraphics[width=0.3\textwidth,height=4cm]{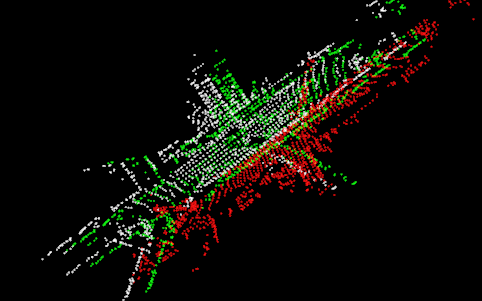}} \hfil
	\subfloat[]{\includegraphics[width=0.3\textwidth,height=4cm]{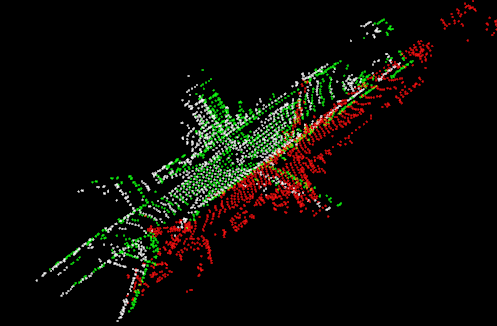}} \hfil
	\subfloat[]{\includegraphics[width=0.3\textwidth,height=4cm]{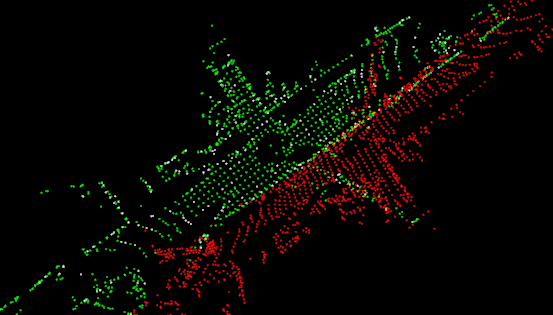}} \hfil
      \caption{(From KITTI dataset) White point cloud: Original Point Cloud(\textit{Target}). Red point cloud: \textit{Source} point cloud and Green point cloud: \textit{Source} transformed point cloud after applying different registration methods. Source is transformed from the \textit{Target} as ($(r,p,y,x,y,z)=(-4.27108,-0.505914,0.0988647,10.938, -10.532, 17.832)$). (a)-(c) shows the convergence of the \textit{Source} point clouds after 7, 15 and 33 iterations using the standard ICP,  respectively (RMSE after 33 iterations is 13.7695). (d)-(f) shows the same using ICP Point to Plane (RMSE after 33 iterations is 12.6385). (g)-(i) shows the same using CoSM ICP (RMSE after 25 iterations is 1.01689e-08).  }
  \label{fig:Res7_lidataset1}
\vspace{-0pt}
\end{figure*}

\begin{table}

\caption{RMSE comparison of different methods with different values of rotation and translation (from \textit{Source} to \textit{Target})after 50 iterations on the KITTI Lidar dataset. Note: Rotation component is in radians}
\scalebox{0.7}[0.7]{%
\begin{tabular}{ |c|c|c|c|c|c|c| } 
\hline
  Transformation(r,p,y,x,y,z) & ICP Standard SVD & ICP Point to Plane & GICP & ICP Non-Linear & NDT & CoSM ICP\\
\hline
(4.23443,-5.64022,-3.13665,-2.2485,-1.19326,-9.12248) & 16.1754 & 371.296 & 38.1373 & 16.1762 & 274.547 & 3.4779e-08\\
(3.64853,-3.29044,-5.13981,0.991908,-5.06315,4.75733) & 16.9795 & 480.78 & 321.312 & 23.8551 & 432.754 & 3.20148e-08\\
(-5.78829,0.576843,1.91183,0.640607,6.7615,3.82172) & 194.469 & 513.454 & 225.474 & 194.581 & 257.164 &  4.47064e-08\\
(-1.0146,-0.427581,-3.42779,7.22804,-7.6078,-0.806072) & 10.2524 & 286.673 & 18.9515 & 10.266 & 24.5575 & 2.83469e-08\\
(0.149154,4.33846,1.82095,-2.5633,-0.377752,-1.20834) & 16.8181 & 1010.43 & 248.244 & 17.2776 & 258.734 & 4.46591e-08\\
\hline
\end{tabular}}
\label{Tab:RMSE_Compare4}
\end{table}

\subsection{Evaluation on datasets with outliers.} 
In this section we evaluated our approach in cases where the data has outliers. We randomly injected data values at random indexes in the \textit{Source} point set. The algorithm for injecting random outliers is shown in Algorithm \ref{outliers_add}.

\begin{algorithm}[!htb]
\begin{algorithmic}[1]
\Function{\textbf{InjectOutliers}}{$\bold{P_s}$}.\Comment{Read the Source point cloud data.}
    \State $ridx$= Pick random index in the Source point set.
    \State $tr_{tx}$= Random translation along the X-axis.
    \State $tr_{ty}$= Random translation along the Y-axis.
    \State $tr_{tz}$= Random translation along the Z-axis.
    \State $\bold{P_s}[ridx].x=\bold{P_s}[ridx].x+tr_{tx}$
    \State $\bold{P_s}[ridx].y=\bold{P_s}[ridx].y+tr_{ty}$
    \State $\bold{P_s}[ridx].z=\bold{P_s}[ridx].z+tr_{tz}$
   
    \State \Return {$\textbf{P}_{s}$} \Comment{Return the transformed point cloud and name it as Source.}

\EndFunction
\end{algorithmic}
\caption{Transform the point cloud with a Random Transformation Matrix.}
\label{outliers_add}

\end{algorithm}
In this experiment we selected the percentage of points in the \textit{Source} dataset to be affected by outliers. We added the outliers after performing voxel grid filtering of leaf size $0.005$ since one can still encounter outliers after pre-processing the data. The \textit{Source} point cloud is transformed from the \textit{Target} point cloud with the transformation $(r,p,y,x,y,z)=(-1.32811,-5.87854,2.12814,-0.874,-0.433,0.221)$. The \textit{Source} point cloud is indeed affected by outliers, and we compare the results with different approaches on different datasets when 10\% ,25\% and 50\% of the \textit{Source} data points are affected.

We compared our approach with other methods as well. Table~\ref{Tab:RMSE_Compare5} shows the RMSE comparison with other datasets where 10\% of the \textit{Source} dataset (after voxel grid filtering) is is affected. Clearly, our methods outperforms other approaches under various rotations and translations. 
Our results are consistent and keeps performing better even when 25\% and 50\% of the \textit{Source} dataset are affected as shown in \ref{Tab:RMSE_Compare6} and \ref{Tab:RMSE_Compare7}.

\begin{figure*}[!htb]
 \centering
    \subfloat[]{\includegraphics[width=0.3\textwidth,height=4cm]{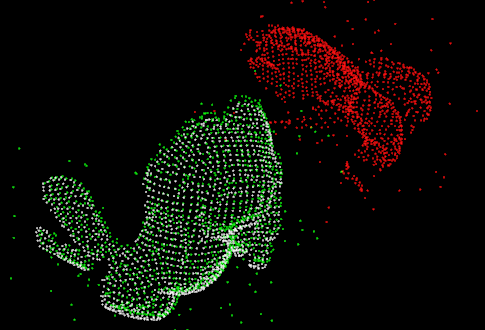}} \hfil
	\subfloat[]{\includegraphics[width=0.3\textwidth,height=4cm]{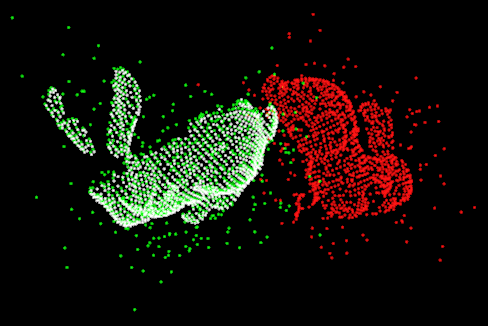}} \hfil
	\subfloat[]{\includegraphics[width=0.3\textwidth,height=4cm]{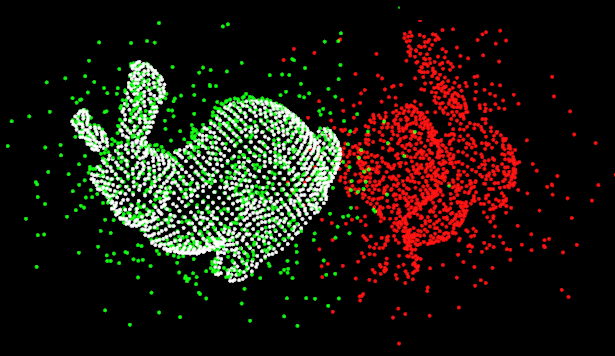}} \hfil
	\subfloat[]{\includegraphics[width=0.3\textwidth,height=4cm]{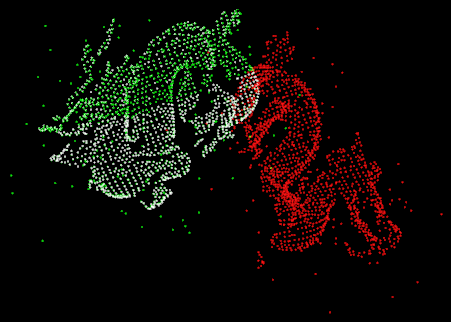}} \hfil
	\subfloat[]{\includegraphics[width=0.3\textwidth,height=4cm]{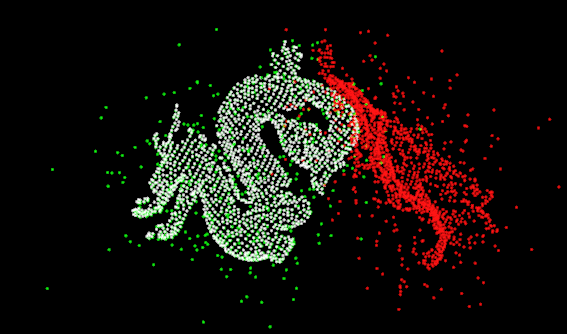}} \hfil
	\subfloat[]{\includegraphics[width=0.3\textwidth,height=4cm]{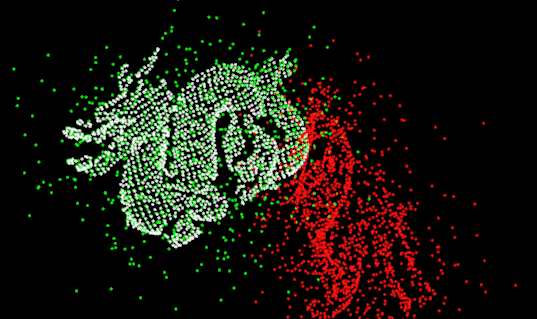}} \hfil
	\subfloat[]{\includegraphics[width=0.3\textwidth,height=4cm]{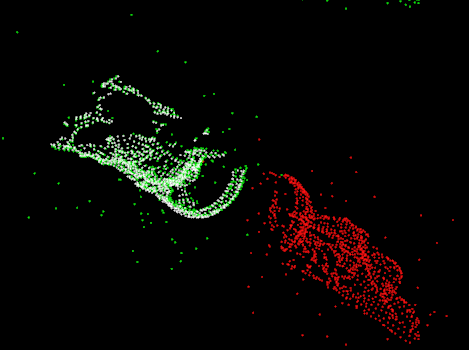}} \hfil
	\subfloat[]{\includegraphics[width=0.3\textwidth,height=4cm]{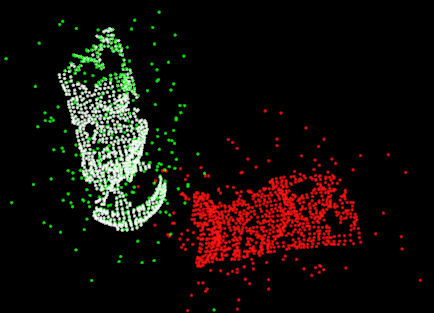}} \hfil
	\subfloat[]{\includegraphics[width=0.3\textwidth,height=4cm]{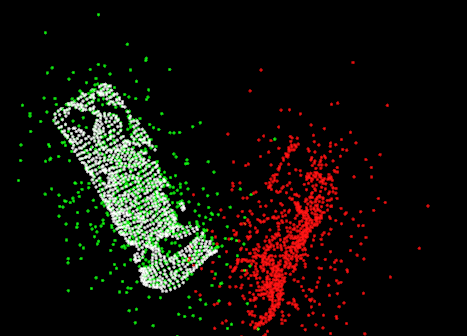}} \hfil
      \caption{CoSM Results when \textit{Source} is infected with noise. White point cloud: Original Point Cloud (\textit{Target}). Red point cloud: Infected \textit{Source} point cloud. Green point cloud: \textit{Source} transformed point cloud after applying iterations. Transformation from \textit{Source} to \textit{Target}: ($(r,p,y,x,y,z)=(1.6585448,0.6090677,-1.204856 ,0.00,-0.003,0.003)$).
      (a),(b) and (c) shows the result of CoSM ICP on Bunny dataset when 10\%, 25\% and 50\%of the data is affected with outliers. Their respective RMSE's are $8.64002e-05$,$0.000203088$ and $0.000384025$. (d),(e) and (f) shows the result of CoSM ICP on the Dragon dataset when 10\%, 25\% and 50\%of the data is affected with outliers. Their respective RMSE's are $6.91792e-05$,$0.000163833$ and $0.000368516$. (g),(h) and (i) shows the result of CoSM ICP on the Happy Buddha dataset when 10\%, 25\% and 50\%of the data is affected with outliers. Their respective RMSE's are $9.29047e-05$,$0.000205528$ and $0.000386966$. For each case we performed around 30 iterations.}
     
  \label{fig:Res7_lidataset1}
\vspace{-0pt}
\end{figure*}

\begin{table}[!htb]
\caption{(10\% of the \textit{Source} data contains outliers) RMSE comparison of different methods with different values of rotation and translation (from \textit{Source} to \textit{Target}) after 50 iterations on different datasets. Note: Rotation component is in radians}
\scalebox{0.7}[0.7]{%
\begin{tabular}{ |c|c|c|c|c|c|c| } 
\hline 
\multicolumn{7}{|c|}{Bunny Dataset}\\
\hline
  Transformation(r,p,y,x,y,z) & ICP Standard SVD & ICP Point to Plane & GICP & ICP Non-Linear & NDT & CoSM ICP\\
\hline
(-2.36822,-0.726873,-5.69612,7.08856,-2.68581,7.63531) & 0.000411328 & 711.375 & 114.581 & 0.000269638 & 114.581 & 6.15845e-05\\
(-2.67216,3.73756,-0.902484,-9.27928,-4.0282,4.09878) & 0.000257389 &  374.895 & 0.000460177 & 23.8551 & 119.48 & 7.70252e-05\\
(-1.56059,4.39878,-2.80893,-3.94166,-3.10092,-3.34176) & 0.000527419 & 304.884 & 37.13 & 0.000487805 & 37.13 &  8.26112e-05\\

\hline
  \multicolumn{7}{|c|}{Dragon Dataset}\\
\hline
(0.177274,-3.62568,5.47677,0.852604,-5.3915,-9.91804) & 0.000220957 & 76.301 & 128.455 & 0.000246111 & 128.455 & 6.49243e-05\\
(-0.620232,-5.82563,3.78543,4.85066,2.74563,-8.42101) & 0.000270745 &  45.484 & 101.049 & 23.8551 & 101.049 & 7.94277e-05\\
(-1.53521,4.48718,-0.388112,-7.74233,5.42215,-6.44992) & 0.00021795 & 241.345 & 125.922  &  0.000223478  & 125.922  &   7.67165e-05\\

\hline
  \multicolumn{7}{|c|}{Happy Buddha Dataset}\\
\hline
(-2.83367,-1.24896,2.98587,2.69059,3.34046,-4.41227) & 0.000176202 & 19.1229 & 36.8922 & 0.000212469 & 36.8922 & 7.10533e-05\\
(-2.47527,-5.65213,-3.30425,-3.8167,8.95128,-9.03962) & 0.000210153 &  598.839 & 174.049 & 0.000217757 & 174.049 & 8.03656e-05\\
(3.05982,-0.897597,-1.43556,1.60636,1.22125,7.01582) & 0.000171509  & 146.972 & 52.0521  &  0.000202291  & 52.0521  &   7.24014e-05\\

\hline
\end{tabular}}
\label{Tab:RMSE_Compare5}
\end{table}

\begin{table}[!htb]
\caption{(25\% of the \textit{Source} data contains outliers) RMSE comparison of different methods with different values of rotation and translation (from \textit{Source} to \textit{Target}) after 50 iterations on different datasets. Note: Rotation component is in radians}
\scalebox{0.7}[0.7]{%
\begin{tabular}{ |c|c|c|c|c|c|c| } 
\hline 
\multicolumn{7}{|c|}{Bunny Dataset}\\
\hline
  Transformation(r,p,y,x,y,z) & ICP Standard SVD & ICP Point to Plane & GICP & ICP Non-Linear & NDT & CoSM ICP\\
\hline
(1.55693,-1.11205,4.17676,-6.05252,1.66971,5.82581) & 0.000426784 & 1175.52 & 71.583 &  0.000365769 & 71.583 & 0.000187812\\
(5.65505,4.15868,-3.27751,9.31612,-5.23313,-5.55587) & 0.000346638 & 3286.84  & 0.000410574  & 144.696 & 144.696 & 0.000167037\\
(-2.55584,-4.76118,3.16486,-7.20725,3.99654,-7.30791) & 0.000447408 & 1343.79 & 117.165 & 0.000552115 & 117.165 &   0.000216027\\

\hline
  \multicolumn{7}{|c|}{Dragon Dataset}\\
\hline
(6.25758,2.3638,-2.88282,-7.36544,-6.20769,-7.92445) & 0.000294534 & 67.9046 & 156.303 & 0.000286822 & 156.303 & 0.000168995\\
(3.76055,2.28707,4.29734,-8.8644,-6.40219,1.19919) &0.000255092 &  115.102 & 120.245 & 0.000292728 & 120.245 & 0.000158364\\
(4.63246,-4.08457,-1.65193,-9.97667,-4.37054,1.77417) & 0.000354324 & 174.437 & 119.75  & 0.000263999   & 119.75  &  0.000162741\\

\hline
  \multicolumn{7}{|c|}{Happy Buddha Dataset}\\
\hline
(-1.20164,-2.47146,4.2161,1.29076,7.69533,-6.64823) & 0.000295019 & 116.015 & 98.0711 & 0.000298207  & 98.0711 & 0.000194422\\
(-1.14973,-4.81181,3.44981,9.95025,9.69451,8.84972) & 0.000266775 &  103.766 & 269.094  & 0.000262239  & 269.094  & 0.000177326\\
(-2.78376,3.84551,-5.11847,-0.469235,-7.93148,-7.98029) & 0.000269701  & 116.321 & 127.049  &   0.000269828  & 127.049 &    0.0001856\\

\hline
\end{tabular}}
\label{Tab:RMSE_Compare6}
\end{table}

\begin{table}[!htb]
\caption{(50\% of the \textit{Source} data contains outliers) RMSE comparison of different methods with different values of rotation and translation (from \textit{Source} to \textit{Target}) after 50 iterations on different datasets. Note: Rotation component is in radians}
\scalebox{0.7}[0.7]{%
\begin{tabular}{ |c|c|c|c|c|c|c| } 
\hline 
\multicolumn{7}{|c|}{Bunny Dataset}\\
\hline
  Transformation(r,p,y,x,y,z) & ICP Standard SVD & ICP Point to Plane & GICP & ICP Non-Linear & NDT & CoSM ICP\\
\hline
(4.21267,-5.1913,0.510754,4.78028,-3.81009,-8.99842) & 0.000555246 & 176.793 & 118.74 &  0.000705962 & 118.74 & 0.000367226\\
(-1.90501,2.94602,-4.15732,-4.38121,-8.43926,0.951326) & 0.000533201 & 1644.03  & 91.374  & 0.000535398 & 91.374 & 0.000364791\\
(-3.88617,5.23854,-4.81381,2.95017,-6.77983,-2.26068) & 0.000615907 & 83.6865 & 61.5916 & 0.000610313 & 61.5916 &   0.00036124\\

\hline
  \multicolumn{7}{|c|}{Dragon Dataset}\\
\hline
(-4.41725,-2.40238,-0.642351,3.74842,-9.32348,-9.14091) & 0.000487516 & 188.167 & 185.885 & 0.000510906 & 185.885 & 0.000372615\\
(3.29757,5.37519,4.68352,5.59739,-1.76298,-7.59656) & 0.000362955  & 487.439  & 90.1547 & 0.000516225 & 90.1547 & 0.000327693\\
(-2.27986,-2.66004,-5.61896,-9.89552,-2.83092,4.9319) & 0.000520534 & 91.5809 & 127.933  & 0.000472107  & 127.933 &  0.000380396 \\

\hline
  \multicolumn{7}{|c|}{Happy Buddha Dataset}\\
\hline
(5.4771,1.17243,-0.67022,-8.42324,-5.29389,-3.70151) & 0.000462191 & 510.054  & 111.689 & 0.000489964  & 111.689 & 0.000384029\\
(1.77436,-2.70017,-4.70393,1.9514,4.74639,-8.0765) & 0.000478957 &  128.078 & 90.493 & 0.000489046 & 90.493  & 0.000383038 \\
(-3.01961,2.91297,-5.51066,6.01917,4.29907,5.78239) & 0.000486108  & 155.204 & 86.8654  &   0.00173299  & 86.8654 &  0.000399139\\

\hline
\end{tabular}}
\label{Tab:RMSE_Compare7}
\end{table}

\subsection{Effect of $\sigma$.}
As shown in the previous section, the registration works well when the value of $\sigma$ is 100. With lower values of $\sigma$ like 0.01 it aligned well for small rotation and translation, however for large rotation and translation between the \textit{Source} and the \textit{Target} it took a large number of iterations to converge. For even smaller values of $\sigma$ like 0.001, and for larger rotation and translation, alignment did take place. Moreover, the iterations was larger ($\sim500$), and the computed transformation was prone to errors. 

\section{Discussion} 
\begin{figure*}[!htb]
 \centering
    \subfloat[]{\includegraphics[width=0.49\columnwidth,height=5cm]{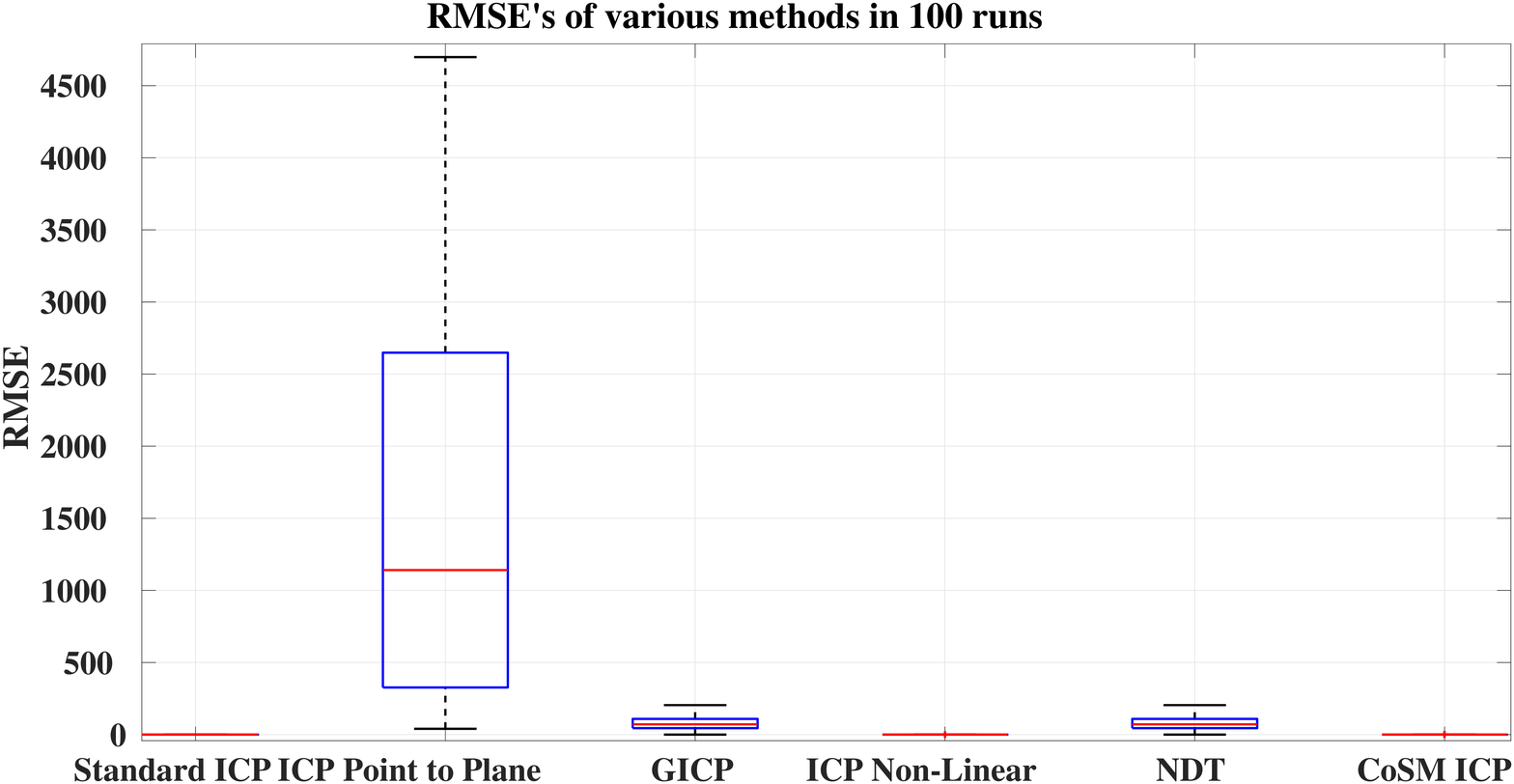}} \
	\subfloat[]{\includegraphics[width=0.49\columnwidth,height=5cm]{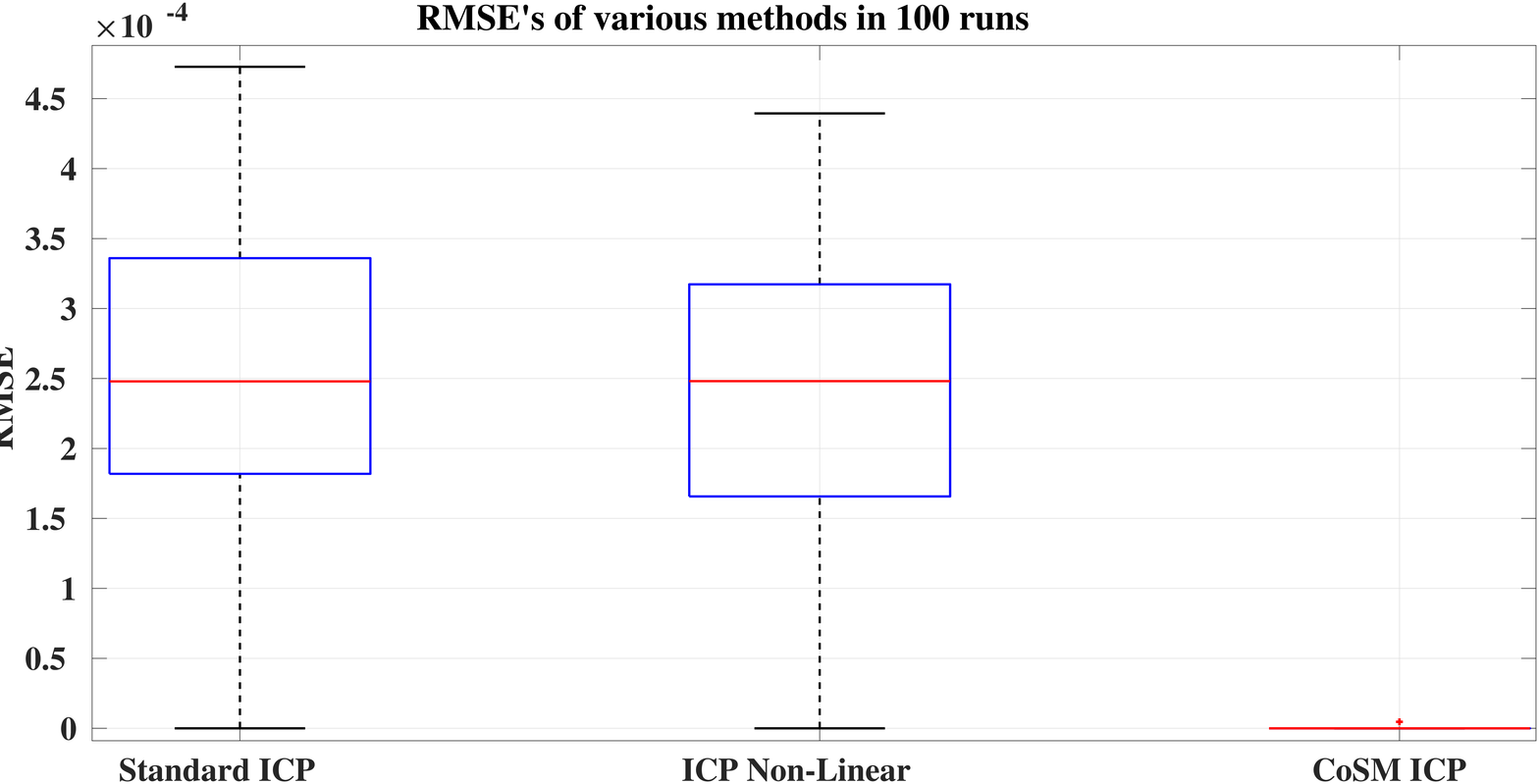}} 
	\caption{Average RMSE's of various methods in 100 runs (Bunny Rabbit dataset). Each run consists of random rotation and translation between the \textit{Source} and the \textit{Target}.  }
  \label{fig:Res8_RMSE100_bun}
\vspace{-0pt}
\end{figure*}

\begin{figure*}[!htb]
 \centering
    \subfloat[]{\includegraphics[width=0.49\columnwidth,height=5cm]{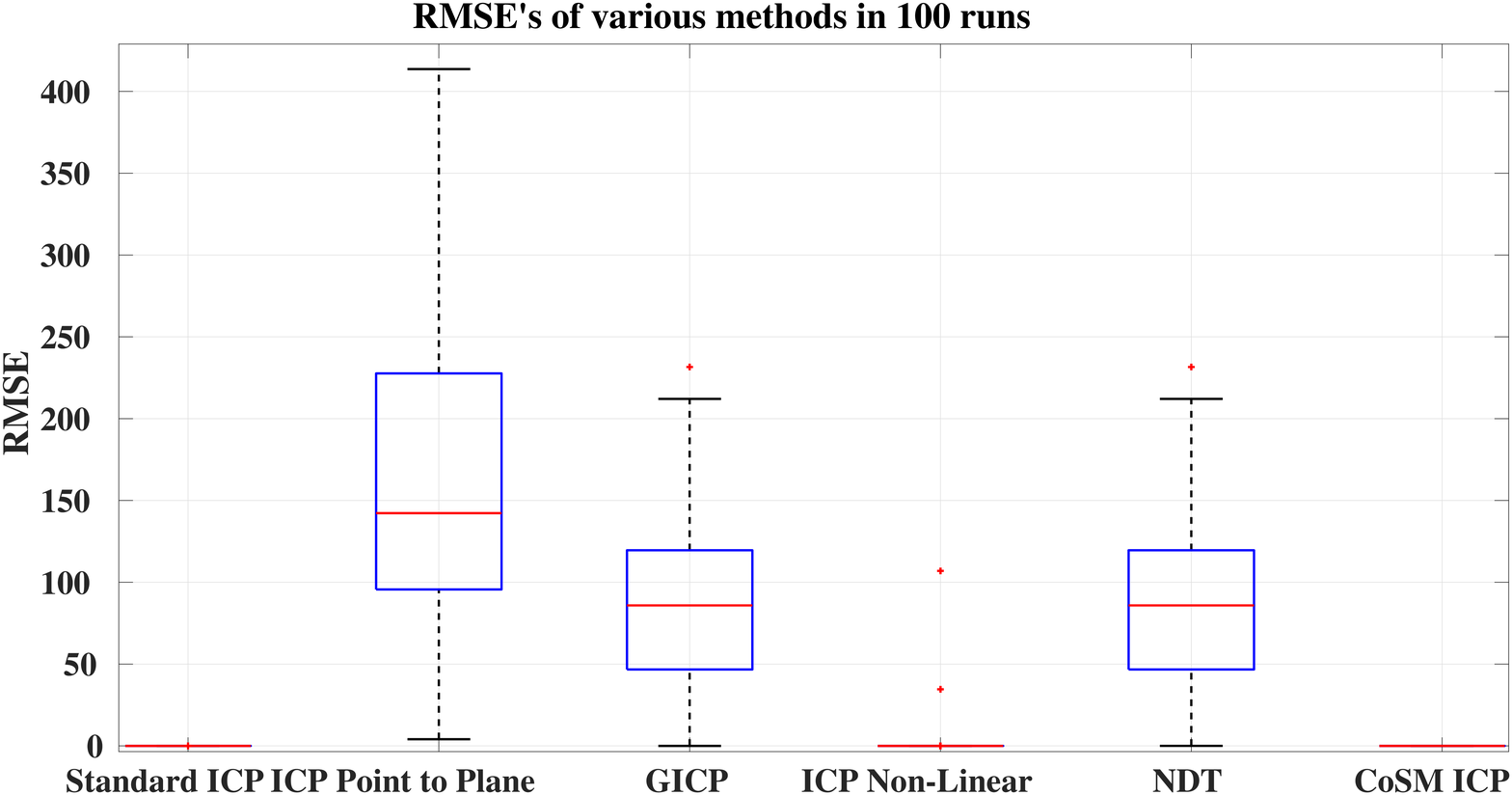}} \
	\subfloat[]{\includegraphics[width=0.49\columnwidth,height=5cm]{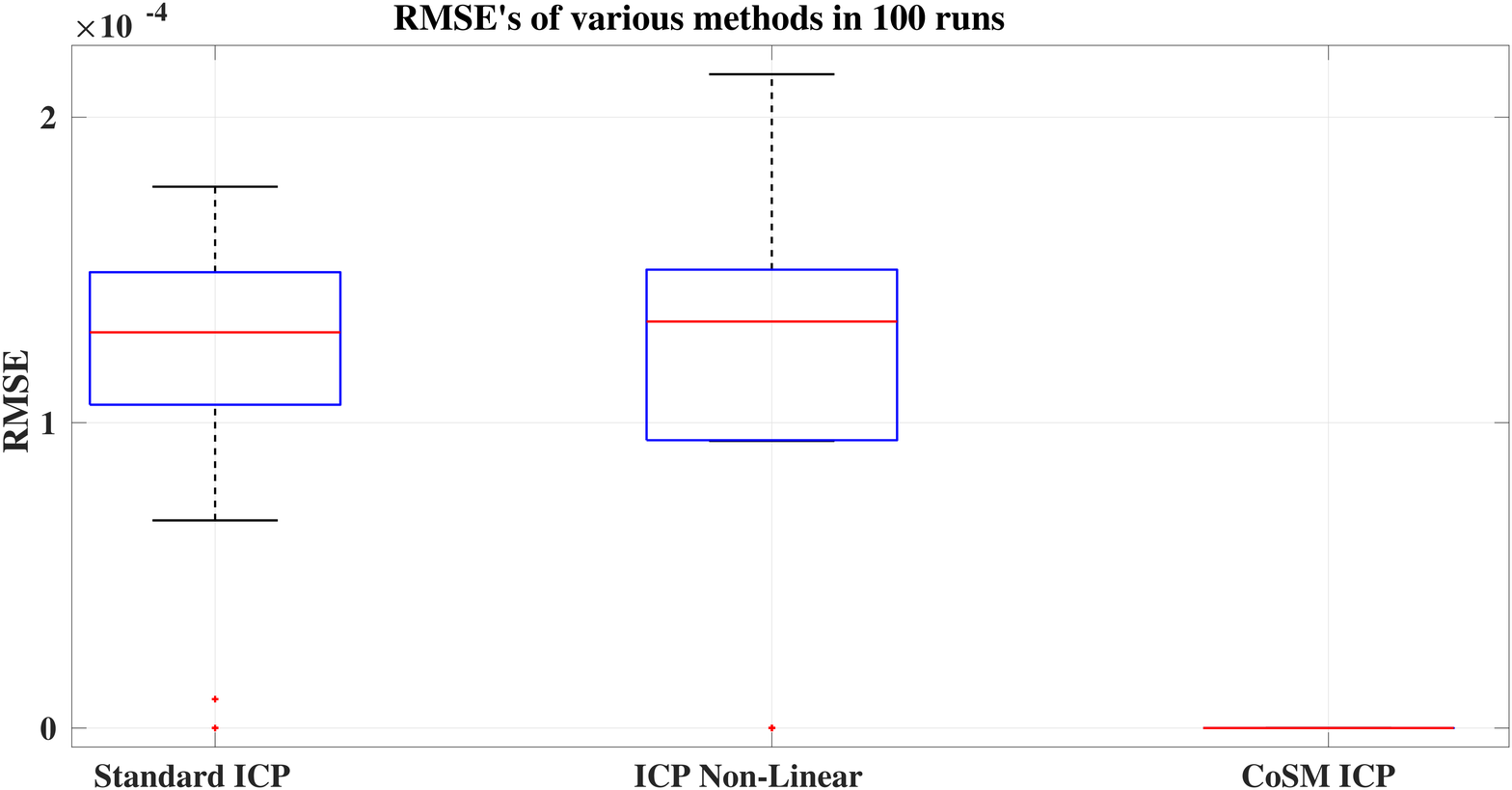}} 
	\caption{Average RMSE's of various methods in 100 runs (Dragon dataset). Each run consist of random rotation and translation between the \textit{Source} and the \textit{Target}. }
  \label{fig:Res9_RMSE100_dr}
\vspace{-0pt}
\end{figure*}

\begin{figure*}[!htb]
 \centering
    \subfloat[]{\includegraphics[width=0.49\columnwidth,height=5cm]{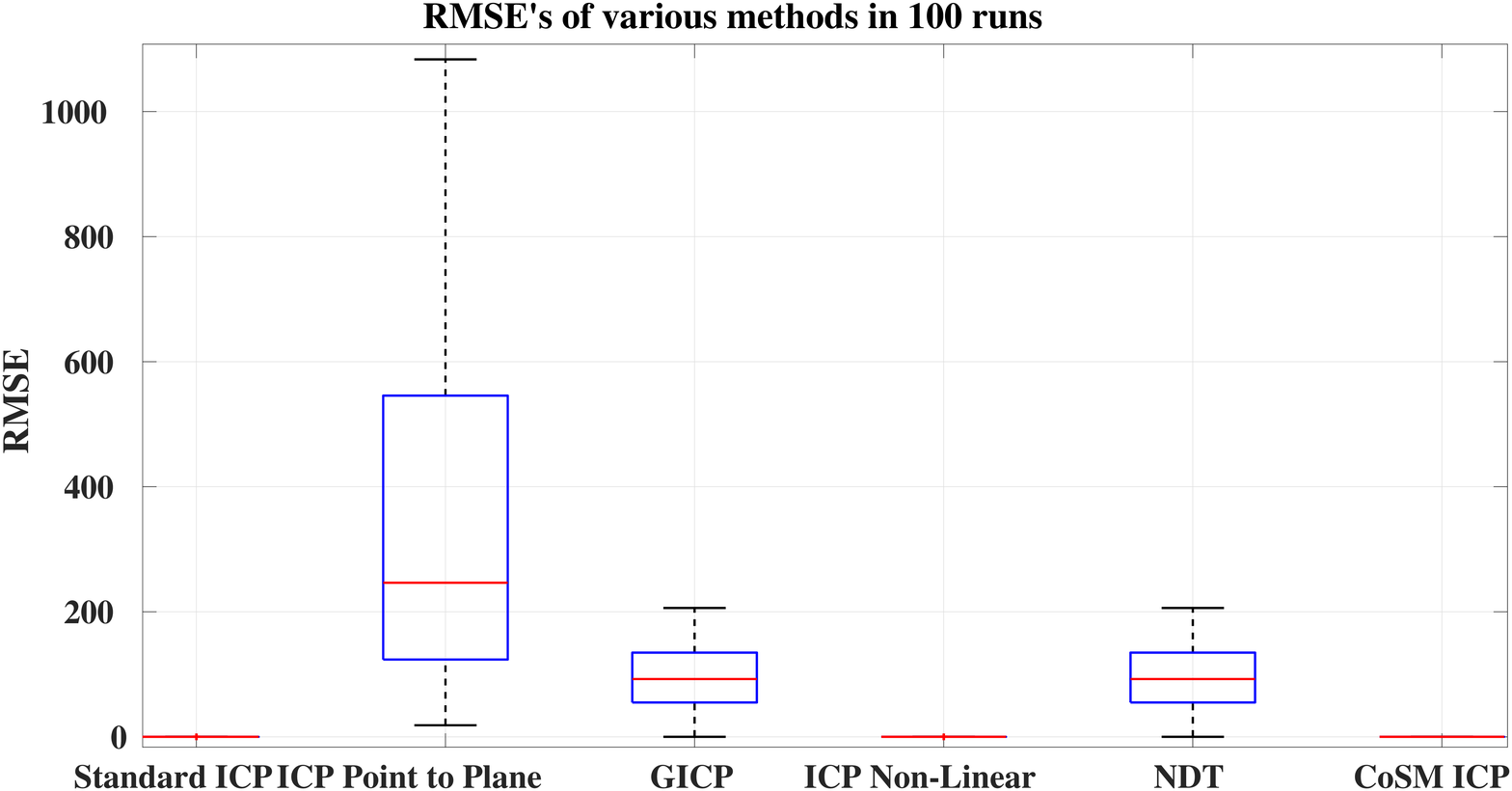}} \
	\subfloat[]{\includegraphics[width=0.49\columnwidth,height=5cm]{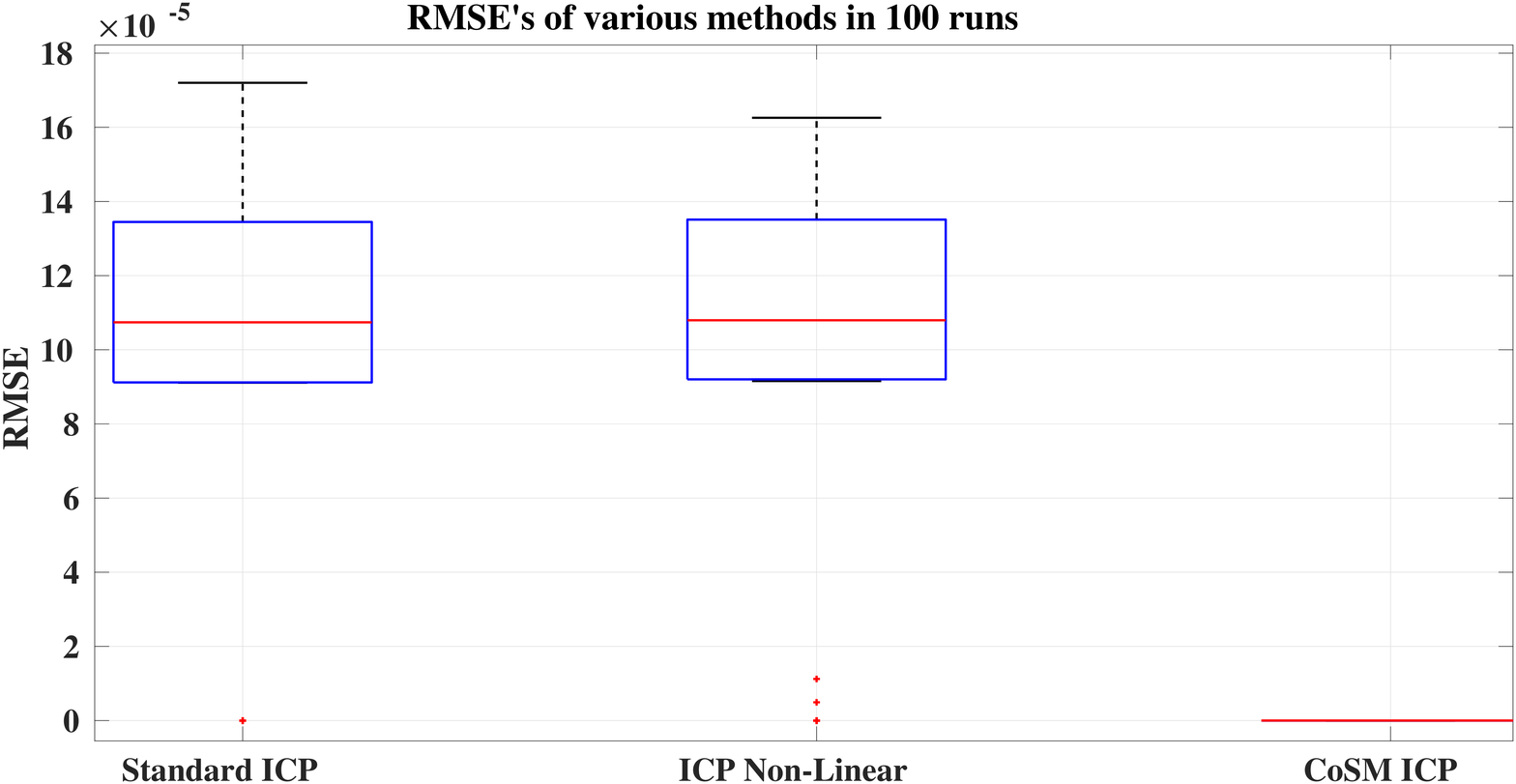}} 
	\caption{Average RMSE's of different methods in 100 runs (Happy Buddha dataset). Each run consist of random rotation and translation between the \textit{Source} and the \textit{Target}. }
  \label{fig:Res10_RMSE100_hp}
\vspace{-0pt}
\end{figure*}

We have found that using a Correntropy Relationship matrix proved to be very effective in the problem of registration. Through experiments we have evaluated that our approach is not only robust to large rotation and translation, but also robust to outliers. We have compared our approach to other state of the art methods like GICP, NDT, ICP Non-Linear etc. and proved it's efficiency under large transformation matrices. It is quite important to note that the size of the Correntropy Matrix is dependent on the size of the dataset, which means that if we have larger datasets, our approach would be computationally very expensive for each iteration ($\sim 7 ms$ for a datset of size 10000 points for each iteration). In addition to our previously mentioned results, we have also evaluated on different datasets using multiple runs. We define a run such that in each run, we generate a random transformation matrix and we compute a transformation matrix from randomly generated translation and rotation components and transform the original point cloud (which we call it as \textit{Target}) to another point cloud (which we call it as \textit{Source}). We then employ several state of the art methods like GICP, NDT , ICP Non-Linear NDT and CoSM ICP iteratively on the \textit{Source} point cloud. In this case we perform the iteration 50 times during each run and collect the final RMSE's from each of the methods. The components of the random transformation matrix (translation and rotations ($x,y,z,r,p,y$)) are generated independently with different seeds (seeds are defined by the system's clock). The mean for all the components is zero. The standard deviation for the translation components along $x,y,z$ is 10, where as for the angular components it is 6.28317 radians (360 degrees) along each axes. This guarantees that we could potentially cover all the possible orientation along all axes.  On the Bunny dataset we found that ICP Point to Plane had a very large RMSE ( mean RMSE for ICP Point to Plane is $1.9668e+03$ in 100 runs) which means the \textit{Source} dataset was not at all aligned to the \textit{Target} dataset at all. However GICP and NDT better than ICP Point to Plane in most of the cases (mean RMSE of GICP in 100 runs is $ 75.6727$ and the mean RMSE of NDT in 100 runs is $ 76.2536$). However in certain transformations GICP and NDT performs well with RMSE of $8.8582e-04$ and $5.6122e-04$ respectively. On the other hand if we compare the average RMSE of Standard ICP, ICP Non-linear and CoSM ICP we see that their average RMSE's in 100 runs is $2.4156e-04$, $2.3781e-04$ and $3.2489e-07$ respectively. It is very evident that CoSM ICP is better than the other approaches (by more than $\sim100\%$). Fig~\ref{fig:Res8_RMSE100_bun}(a) shows the distribution of errors (RMSE's) of various methods in 100 runs on the Bunny Rabbit dataset. Fig. \ref{fig:Res8_RMSE100_bun}(b) shows the RMSE deviation of Standard ICP, ICP Non-Linear and CoSM ICP. In the Dragon dataset the average RMSE's of Standard ICP, ICP Point to Plane, GICP, ICP Non-Linear, NDT and CoSM ICP is $1.1372e-04$, $177.0397$, $86.8183$, $1.4154$, $87.5327$ and $2.6989e-13$ respectively. In this case as well CoSM outperformed the other approaches by more than $\sim100\%$. It is again evident from Fig. \ref{fig:Res9_RMSE100_dr}(a) and  Fig. \ref{fig:Res9_RMSE100_dr}(b). On the Happy buddha dataset, the average RMSE's for Standard ICP, ICP Point to Plane, GICP, ICP Non-Linear, NDT and CoSM ICP is $ 9.4830e-05$, $414.0633$, $93.0712$, $1.1289e-04$, $93.3621$ and $5.6055e-13$ respectively in 100 runs. Again it is validated that CoSM ICP performs better than the rest of the approaches by more than $\sim100\%$. Fig. \ref{fig:Res10_RMSE100_hp}(a) and Fig. \ref{fig:Res10_RMSE100_hp}(b) clearly validates our claim for the happy buddha dataset. Readers can note that the top and the bottom of the box plot shown represent the $25^{th}$ and $75^{th}$ percentiles respectively and the middle red line shows the median of the error (average RMSE's) across 100 runs. 

It is significant to note that our approach has proved beneficial in solving the registration problem under various rotations and translations as compared to other well-known methods. We encourage the community to validate our approach on their own datasets.

\section{Conclusions and Future Work} 
One of the key things we have addressed is to find an accurate estimate of the transformation between the \textit{Source} and the \textit{Target} point clouds under various rotations and translations. In this work, we have verified our approach in various datasets, where the \textit{Source} is transformed from the \textit{Target} using various randomly generated transformation matrices. Through the obtained results we can see that our method has performed better than the other state of the art approaches. In addition, we also evaluate our approach when the \textit{Source} dataset is affected by noise generated with random intensity and affecting random data points. In this scenario as well, we have verified that our proposed approach has outperformed most of the other state of the art approaches where the `infected' \textit{Source} dataset is transformed randomly from the \textit{Target} datset. We strongly insist our readers to evaluate our approach located in our Github repository. In addition to our current approach, we also aim to evaluate our approach in SLAM where the poses computed from our approach can be used to localize a robot (both aerial and ground robot) as well as build a map of the environment. For further evaluation, we plan to test our approach in the well-known KITTI SLAM dataset and other online datasets.

\section*{Acknowledgement}
This work is partially supported   by the U.S. National Science Foundation (NSF) under grants NSF-CAREER: 1846513 and NSF-PFI-TT: 1919127, and the U.S. Department of Transportation, Office of the Assistant Secretary for Research and Technology (USDOT/OST-R) under Grant No. 69A3551747126 through INSPIRE University Transportation Center.  
The views, opinions, findings and conclusions reflected in this publication are solely those of the authors and do not represent the official policy or position of the NSF and USDOT/OST-R.





\bibliographystyle{elsarticle-num}
\bibliography{main_PR_Elsvier.bib}







\end{document}